\documentclass[10pt,journal,compsoc]{IEEEtran}
\usepackage{amsmath,amssymb,amsfonts}
\usepackage{booktabs}
\usepackage{algorithm}
\usepackage[noend]{algpseudocode}
\usepackage{times}
\usepackage{footnote}
\usepackage{multirow}
\usepackage{subcaption}
\usepackage{textcomp}
\usepackage{url}
\usepackage{verbatim}
\usepackage{graphicx}
\usepackage{wrapfig}
\usepackage{xcolor, colortbl}
\usepackage{tikz}

\definecolor{ForestGreen}{RGB}{34,139,34}

\definecolor{yellow}{RGB}{238, 218, 172}
\definecolor{lightgreen}{RGB}{194, 218, 194}

\definecolor{codegray}{gray}{0.95}
\definecolor{codeblue}{rgb}{0.0,0.2,0.6}

\ifCLASSOPTIONcompsoc

  \usepackage[nocompress]{cite}
\else
  \usepackage{cite}
\fi
\ifCLASSINFOpdf

\else
\fi

\begin{document}

\title{Multi-Domain Learning with\\ Global Expert Mapping}

\author{Pourya~Shamsolmoali,~\IEEEmembership{Senior Member,~IEEE,}
        Masoumeh~Zareapoor,~\IEEEmembership{Member,~IEEE,} Huiyu~Zhou,
        Oscar Mendez,~\IEEEmembership{Member,~IEEE,}
        Dacheng~Tao,~\IEEEmembership{Fellow,~IEEE},
        and Xuelong~Li,~\IEEEmembership{Fellow,~IEEE}

\thanks{P.~Shamsolmoali and M.~Zareapoor are with the Department of Computer Science, University of York, UK (\{pshams55, mzarea222\}@gmail.com)}
\thanks{H.~Zhou is with the School of Computing and Mathematical Sciences, University of Leicester, UK (hz143@leicester.ac.uk).}
\thanks{O.~Mendez is with Centre for Vision Speech and Signal Processing, University of Surrey, UK (o.mendez@surrey.ac.uk).}
\thanks{D.~Tao is with the College of Computing and Data Science at Nanyang Technological University, Singapore (dacheng.tao@gmail.com).}
\thanks{X.~Li is with the Institute of Artificial Intelligence (TeleAI) of China Telecom (xuelong\_li@ieee.org).}
}

\IEEEtitleabstractindextext{%
\begin{abstract}
Human perception generalizes well across different domains, but most vision models struggle beyond their training data. This gap motivates multi-dataset learning, where a single model is trained on diverse datasets to improve robustness under domain shifts. However, unified training remains challenging due to inconsistencies in data distributions and label semantics. Mixture-of-Experts (MoE) models provide a scalable solution by routing inputs to specialized subnetworks (experts). Yet, existing MoEs often fail to specialize effectively, as their load-balancing mechanisms enforce uniform input distribution across experts. This fairness conflicts with domain-aware routing, causing experts to learn redundant representations, and reducing performance especially on rare or out-of-distribution domains. We propose GEM (Global Expert Mapping), a planner-compiler framework that replaces the learned router with a global scheduler. Our planner, based on linear programming relaxation, computes a fractional assignment of datasets to experts, while the compiler applies hierarchical rounding to convert this soft plan into a deterministic, capacity-aware mapping. Unlike prior MoEs, GEM avoids balancing loss, resolves the conflict between fairness and specialization, and produces interpretable routing. Experiments show that GEM-DINO achieves state-of-the-art performance on the UODB benchmark, with notable gains on underrepresented datasets and solves task interference in few-shot adaptation scenarios.
\end{abstract}

%%%Training a unified model across diverse datasets is essential for real-world perception, yet differences in data distributions, annotation conventions, and label spaces cause severe optimization instability and systematic underrepresentation of minority domains. Mixture-of-Experts (MoE) architectures offer an attractive solution through input-dependent routing to specialized experts. 

%%%Modern perception systems require learning from multiple, diverse datasets to achieve robust generalization. When we train a single model on these heterogeneous domains, gaps in data distributions, and annotation styles destabilize optimization, leave minority domains underrepresented. Mixture-of-Experts (MoE) models offer a promising solution by routing each input to a specialized expert. Yet, they fail to achieve meaningful domain specialization due to reliance on load-balancing mechanisms that enforce uniform input distribution across all experts. This fairness conflicts with domain-aware routing, causing experts to learn redundant representations, and degrading performance especially on rare or out-of-distribution domains.

}

\maketitle

\IEEEdisplaynontitleabstractindextext

\IEEEpeerreviewmaketitle

\ifCLASSOPTIONcompsoc
\IEEEraisesectionheading{\section{Introduction}\label{sec:introduction}}
\else
\section{Introduction}
\label{sec:introduction}
\fi

\IEEEPARstart{H}{u}man perception naturally generalizes across diverse environments and visual conditions, but most computer vision models are trained on a single dataset, limiting their ability to operate beyond the training distribution \cite{tang2025zero, pan2024large, mi2025learning, li2023sparse}. When applied to new domains with different visual features or label semantics, these models often show significant performance drops \cite{shamsolmoali2024setformer, zhang2025learning}. Multi-dataset learning is therefore essential for building models that are robust to these shifts, yet it introduces fundamental challenges. First, datasets from different domains (e.g., indoor scenes vs. medical images) have disjoint label spaces and divergent visual characteristics, complicating unified training. Second, large datasets like COCO \cite{lin2014microsoft} tend to dominate optimization, while smaller ones like Comic \cite{inoue2018cross} are underrepresented, leading to biased learning. Third, as more datasets are included, domain interference can cause dataset collapse \cite{li2023sparse}, where performance on one domain degrades as the model adapts to another \cite{rebuffi2018efficient, jain2024damex}. Recent methods have proposed various strategies to mitigate these issues. Zhou et al. \cite{zhou2022simple} train separate detectors per dataset and later merge them, avoiding interference during training but suffering post-merge performance loss. Joint training approaches \cite{wang2019towards, meng2023detection, rebuffi2018efficient, kapidis2021multi} improve efficiency, but often require domain knowledge at test time, limiting their use in domain-agnostic settings \cite{shi2024plain}.
%%%%%%%%%%%%%%%%%%%%
A promising alternative is the Mixture-of-Experts (MoE) architecture \cite{shazeer2017sparsely}, which replaces a single dense model with $n$ specialized subnetworks (experts) and uses a gating network to route each input to the most appropriate expert \cite{li2023sparse, jain2024damex, wang2024remoe, zhou2025mergeme, oksuz2024mocae}.   In a multi-domain setting, an ideal MoE would specialize by domain (e.g.,  one expert for Comic data, another for medical), yielding targeted representations per dataset. However, most MoE models fail to achieve this due to reliance on auxiliary load-balancing loss \cite{fedus2022switch}. This loss prevents expert collapse (where one expert handles most inputs) \cite{dai2024deepseekmoe, oldfield2024multilinear}, by encouraging the gating network to distribute tokens uniformly across experts \cite{dai2024deepseekmoe, oldfield2024multilinear}. Yet, this uniform routing directly conflicts with specialization \cite{omi2025load, li2025uni}, penalizing the gate for assigning domain-specific data to a single expert. This leads to (i) knowledge redundancy, where all experts learn similar, mixed-domain representations; and (ii) underrepresented/rare domains receive insufficient focus, as their tokens are scattered across experts \cite{li2023sparse, hayes2024buffer, huang2024harder}. These limitations are especially problematic in high-stakes domains like medical imaging, where clear specialization (e.g., radiology vs. pathology) is critical for interpretability and safety \cite{dai2024deepseekmoe,li2023sparse}.

\begin{figure*}[t]
    \centering
    \begin{minipage}{0.46\textwidth}
        \centering
      \includegraphics[width=\columnwidth]{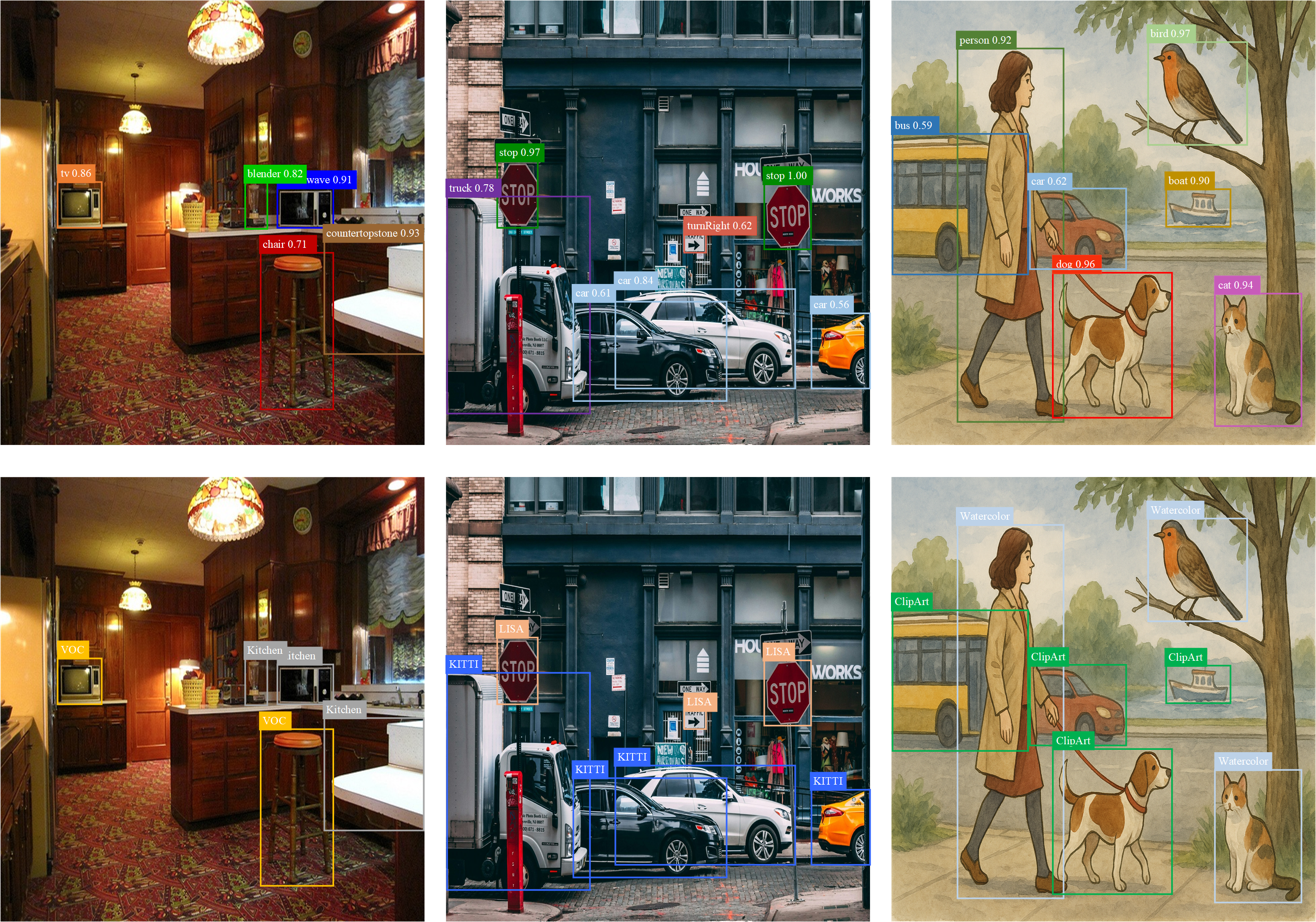} 
    \end{minipage}
   \quad
    \begin{minipage}{0.35\textwidth}
        \centering
        \includegraphics[width=\linewidth]{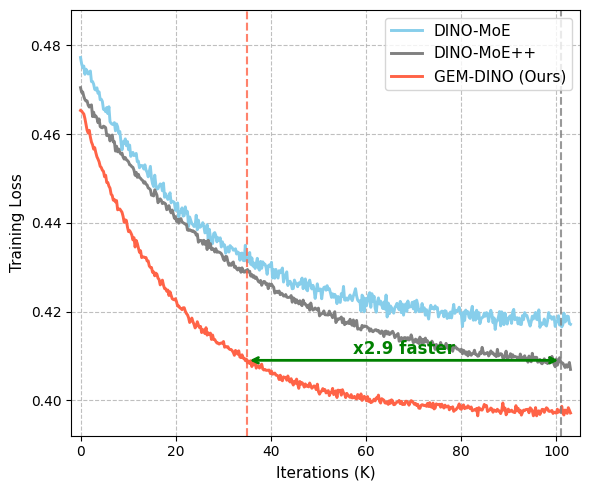}
    \end{minipage}
     %\hfill
      \caption{Improved detection quality and convergence with GEM-DINO. The training loss shows that our GEM-DINO (red) converges $2.9\times$ faster and achieves a lower final loss compared to DINO-MoE and DINO-MoE++, indicating a more stable optimization.} \label{det}
\end{figure*}

To overcome these limitations, we introduce GEM (Global Expert Mapping), a planner-compiler framework inspired by resource scheduling systems \cite{wang2018machine, mosaner2022machine, fischer2020improved}. GEM treats expert routing as a global resource allocation problem. The planner uses linear programming relaxation (LPR) to compute a soft assignment of limited resources (experts) to tasks (datasets). The compiler then applies a hierarchical rounding algorithm \cite{roy2017hierarchical, bhattacharya2023dynamic} to convert this fractional solution into a deterministic, capacity-respecting mapping.  Rounding is performed by decomposing fractional weights into binary components and assigning them iteratively, ensuring that all datasets are routed without exceeding expert capacities. To illustrate this, consider a hospital with $n$ specialist doctors (experts) and $m$ types of patients (datasets). A naive fairness rule ($\mathcal{L}_{\text{balance}}$), might force each doctor to handle an equal mix of patients, resulting in $n$ generalist. The heart surgeon (expert 1) is forced to treat skin rashes (dataset 2). This is what a standard MoE routing does in multi-domain settings. In GEM, the LPR planner first looks at \(m\) patient types and \(n\) doctors expertise (the cost, $w_{ij}$) to draft a soft plan (e.g., 70\% of cardiac cases should go to the heart surgeon). The rounding compiler turns this plan into a determinist schedule that respects the \(c\)-patient-per-doctor limit. This plan-then-compile approach offers four key advantages over standard MoEs: (i) it formulates routing as a global matching problem rather than a per-token classification; (ii) it produces deterministic and interpretable assignment; (iii) it enforces expert capacity explicitly, eliminating auxiliary balance losses; (iv) it enables adaptation to new datasets via re-planning. Prior routing methods like BASE \cite{lewis2021base} and Sinkhorn-MoE \cite{clark2022unified} aim to improve token-expert balance but rely on linear assignment or optimal transport, which are computationally expensive and scale poorly to heterogeneous, multi-dataset settings \cite{liu2022sparsity}. In contrast, GEM is the first to combine LPR-based planning with capacity-aware hierarchical rounding, bridging the gap between optimal scheduling theory and scalable expert routing.
%
%%%%%%%%%%%%%%%%%%%%
We integrate GEM into DINO object detector \cite{zhang2023dino} and evaluate on UODB benchmark \cite{wang2019towards} (see Fig. \ref{det}). GEM outperforms SOTA MoE baselines, achieving +2.1 AP over SoftMoE \cite{puigcerver2023sparse}, +0.8 AP over REMoE \cite{wang2024remoe}, and +1.2 AP over MoE++ \cite{jin2024moe++}. It also surpasses non-MoE baselines by +2.9 and +1.4. On joint training across COCO, Object365, and Visual Genome, GEM improves over DINO-MoE++ by +2.6, +3.9, and +2.0 AP, respectively, showing strong cross-domain generalization. In a further ablation, we apply GEM to Vision-MoE \cite{riquelme2021scaling} for large-scale image classification. Our matching-based routing not only simplifies training, compared to transport-based routing methods \cite{liu2022sparsity, clark2022unified}, but also achieves higher accuracy. %Together, these results highlight GEM’s scalability, robustness, and effectiveness in multi-dataset settings.

%%%%%%%%%%%%%%%%%%%%%%%%%%%%%%%%%%%%%%%% Related-Work section %%%%%%%%%%%%%%%%%%%%%%%%%%%%%%%%%%%%%%%%%%

\vspace{-3pt}

\section{Related Work}
\label{sec:related}
\subsection{Multi-Domain Object Detection (MDOD)}
Object detectors trained on a single dataset, often fail to generalize to new domains due to visual variations such as viewpoint shifts, lighting changes, and out-of-distribution objects \cite{zhang2023dino, liu2024grounding, shamsolmoali2024setformer}. As a result, detectors that perform well in one setting may generalize poorly in another  \cite{zhou2022simple, kapidis2021multi, mi2025learning, lu2024multi, xihan2024generalizing}. A common strategy to overcome this is multi-dataset learning, which trains a single model on a diverse set of datasets at once \cite{girdhar2022omnivore, zhou2022simple, rebuffi2018efficient, shi2024plain}. While this setup improves generalization, it also introduces challenges such as inconsistent label spaces, missing annotations, and severe dataset imbalance \cite{wang2019towards}. Early work addressed this by using a shared backbone with dataset-specific heads \cite{rebuffi2018efficient, yang2019detecting}, allowing the model to separate knowledge across domains. Others used pseudo-labeling and loss reweighting to handle missing labels or partially annotated data \cite{zhou2022simple, shi2024plain}. Wang et al. \cite{wang2019towards} propose squeeze-and-excitation modules to promote cross-domain generalization in a universal detector. More recent approaches introduce unified label spaces through language embeddings \cite{meng2023detection, shi2024plain}. Despite these advances, training imbalance remains a major bottleneck: large-scale datasets (e.g., COCO) tend to dominate optimization, while smaller one (e.g., Comic) are underrepresented. To address this, recent studies have turned to mixture-of-experts architectures \cite{jain2024damex, oksuz2024mocae, li2023sparse, dai2024deepseekmoe}, which route inputs to a subset of experts through a gating mechanism. However, as we discuss next, most MoE-based detectors fail to achieve true domain specialization, especially under dataset imbalance. 

\vspace{-7pt}

\subsection{ Mixture-of-Experts (MoE)}
MoE architectures use a router to assign inputs to specialized experts \cite{shazeer2017sparsely, fedus2022switch}. While dynamic routing (e.g., top-k gating) is commonly used for their adaptability and sparse activation \cite{du2022glam, zhou2022mixture, riquelme2021scaling, krajewski2024scaling}, it suffers from expert collapse, where only a few experts dominate, leaving others underutilized \cite{clark2022unified, hayes2024buffer}. To prevent this, load-balancing mechanisms \cite{fedus2022switch, dai2024deepseekmoe, oldfield2024multilinear} are introduced, which are the primary source of the challenges in multi-domain learning. In this setting, the MoE training objective, $\mathcal{L} = \mathcal{L}_{\text{task}} + \mathcal{L}_{\text{balance}}$, introduces a mathematical contradiction between its two components. The task loss wants specialization; to get the best accuracy, it should assign all data from a specific domain (e.g., medical) to a single expert, which is an inherently non-uniform goal. However, the balancing loss wants uniformity; to prevent expert collapse, it forces the router to scatter all domains across all experts to keep the load even \cite{jain2024damex, li2023sparse}. This conflict (simultaneously specialized and uniform) leads to redundant experts, as all experts are forced to learn a similar, mix of all domains, undermining the purpose of an MoE \cite{li2023sparse, hayes2024buffer, omi2025load}. The literature contains several attempts to solve this conflict, each with its own limitations. Some methods replace the balancing loss by optimal transport \cite{clark2022unified} or bipartite matching \cite{lewis2021base}. While these approaches are more principled, they are computationally expensive and scale poorly for training on heterogeneous datasets \cite{liu2022sparsity}. Static, rule-based \cite{jain2024damex, jiang2024mixtral} or heuristic \cite{akrour2021continuous} routers are fast and simple, but they cannot support fine-grained specialization \cite{li2025uni, oldfield2024multilinear}. Other methods propose architectural changes to the experts themselves. MoE++ \cite{jin2024moe++} introduces a hybrid design with auxiliary experts and gating residuals, but it still relies on the same flawed softmax-based routing logic.  A different line of work, $\mu\text{MoE}$ \cite{oldfield2024multilinear}, addresses expert redundancy by factorizing monolithic experts into pools of smaller factor matrices. This allows virtual experts to be constructed on the fly, reducing parameter redundancy.  Similarly, SIMBAL \cite{omi2025load} reduce redundancy via regularization, but they rely on online adaptive policies that increase training overhead \cite{omi2025load}. REMoE \cite{wang2024remoe} takes a different path, replacing the standard top-k with a ReLU-based router and using an adaptive L1 regularization (instead of $\mathcal{L}_{\text{balance}}$) to enforce both sparsity and load balancing.

\begin{figure}[t]
    \centering
\includegraphics[width=0.90\columnwidth]{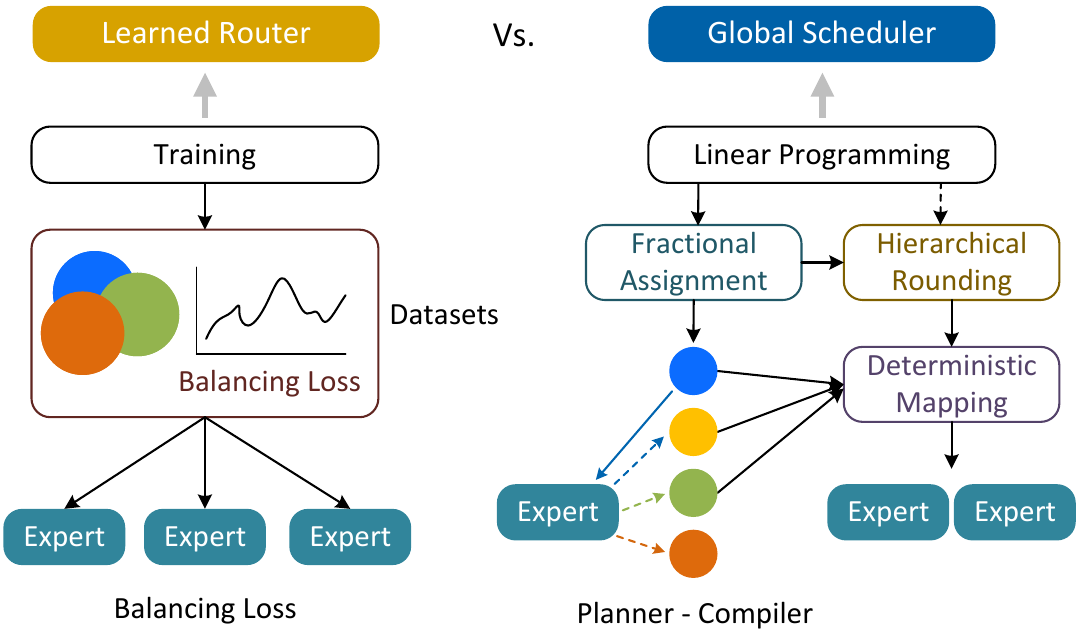}
    \caption{MoE routing vs. our proposed GEM. Traditional MoEs use a learned router with a load-balancing loss, which enforces uniform token distribution and hinders specialization. On Right, GEM replaces the router with a planner-compiler pipeline, enabling deterministic, expert assignment without balancing constraints.}
    \label{pipe}
\end{figure}

Different from these approaches, our GEM (see Fig. \ref{pipe}), introduces a planner-compiler strategy that solves the specialization-vs-uniformity contradiction by decoupling the assignment logic from training. The planner (LPR) computes a fractional plan, globally matching datasets to experts based on specialization-driven costs. The compiler (hierarchical rounding) then converts this soft plan into a deterministic schedule that respects all capacity constraints. This scheduling approach eliminates the need for a balancing loss, allowing the system to achieve true domain specialization.

%%%%%%%%%%%%%%%%%%%%%%%%%%%%%%%%%%%%% Proposed-model section %%%%%%%%%%%%%%%%%%%%%%%%%%%%%%%%%%%%%%%%%

\vspace{-3pt}

\section{Proposed Model}
\label{sec:formatting}
In this section, we begin with describing the problem, then formulate our proposed model and present its architecture. 

\vspace{-5pt}

\subsection{How MoEs Impacts MDOD Performance} 
We consider the multi-domain object detection (MDOD) setting, where the goal is to train a unified model on $m$ diverse datasets $\{D_1, D_2, \ldots, D_m\}$, each with distinct label spaces, dataset sizes, and visual distributions. A promising direction for this setting is the MoE architecture, which uses $n$ expert subnetworks $\{E_1, E_2, \dots, E_n\}$, and a gating function $G(x)\!\in\!\mathbb{R}^n$ to route each input $x$ to a subset of experts. The model output is given by
%%%%%%%%%%%%%%%%%%%%
\begin{equation}
y\!=\!\sum_{j=1}^n G_j(x) \cdot E_j(x), \qquad G_j(x)\ge 0.
\end{equation}
%%%%%%%%%%%%%%%%%%%%
To reduce computation, modern MoEs use sparse top-$k$ routing, activating only a few experts per input \cite{zhou2022mixture}. However, this often leads to expert collapse, where the router overuses a small subset of experts while ignoring others \cite{hayes2024buffer, dai2024deepseekmoe}. To mitigate this, a load-balancing loss \cite{fedus2022switch} is added to the training objective
%%%%%%%%%%%%%%%%%%%%
\begin{equation}
    \mathcal{L}\!=\!\mathbb{E}_{(x,y)\sim\mathcal{D}} [ \ell(y,\sum_j G_j(x) E_j(x))]\;+\;
\lambda\,\mathcal{L}_{\text{balance}}(G),
\end{equation}
%%%%%%%%%%%%%%%%%%%%
where $\ell$ is the task loss, and $\mathcal{L}_{\text{balance}}$ penalizes uneven expert usage. This regularization works well in single-domain settings \cite{oldfield2024multilinear}, but fails in multi-domain training (where inputs come from distinct distributions ${\mathcal{P}_1,\dots,\mathcal{P}_m}$). To understand why, let $P(j \mid D_i)$ denote the probability that the router assigns an input from $D_i$ to expert $E_j$, and let $\alpha_i$ be the sampling probability of $D_i$, which is often highly imbalanced (e.g., $\alpha_{\text{COCO}} \gg \alpha_{\text{Comic}}$). Ideally, the router should assign all inputs from $D_i$ to a dedicated expert $E^*_i$
%%%%%%%%%%%%%%%%%%%%
\begin{equation}
    P(j | D_i) = \begin{cases} 1 & \text{if } j = E^*_i \\ 0 & \text{otherwise} \end{cases}
\end{equation}
%%%%%%%%%%%%%%%%%%%%
This leads to a non-uniform expert usage $P(j) = \sum_i \alpha_i P(j|D_i)$. However, the balancing term $\mathcal{L}_{\text{balance}}$ explicitly penalizes this structure by enforcing $P(j) \approx 1/n$ for all experts $j$. This creates a conflict in the objective, $\min_G (\mathcal{L}_{\text{task}} + \lambda\mathcal{L}_{\text{balance}})$, where the first term encourages specialization (non-uniform $P(j)$), while the second term enforces uniformity. This leads to three major failure modes. (i) Blended experts (mode collapse), where the router assigns mixed-domain inputs to each expert, leading to redundant expert representations \cite{hayes2024buffer, huang2024harder}. (ii) Since $G(x)$ is local and operates on input features, it may routes examples from $D_i$ to experts primarily trained on $D_{j \ne i}$, weakening specialization \cite{chi2022representation}. (iii) The router trained over the full dataset distribution. When a new dataset $D_{m+1}$ is added, the gate must be retrained from scratch over all data, making the system hard to update. To overcome these limitations, we reformulate expert routing as a global dataset-to-expert mapping, $\rho: \{1, \dots, m\} \to \{1, \dots, n\}$, that assigns datasets to experts while satisfying capacity constraints (i.e., how many datasets each expert can support). This is a form of the generalized assignment problem, where the goal is to maximize domain specialization under hard resource limits \cite{roy2017hierarchical, jambulapati2022regularized, fischer2020improved}. We solve this assignment through our planner-compiler strategy.  %This design is particularly useful in high-stakes domains (healthcare or finance), where domain-specific specialization are crucial.

\vspace{-5pt}

\subsection{Global Expert Mapping (GEM)} \label{model}
\begin{figure}[t]
    \centering
    \includegraphics[width=1.02\columnwidth]{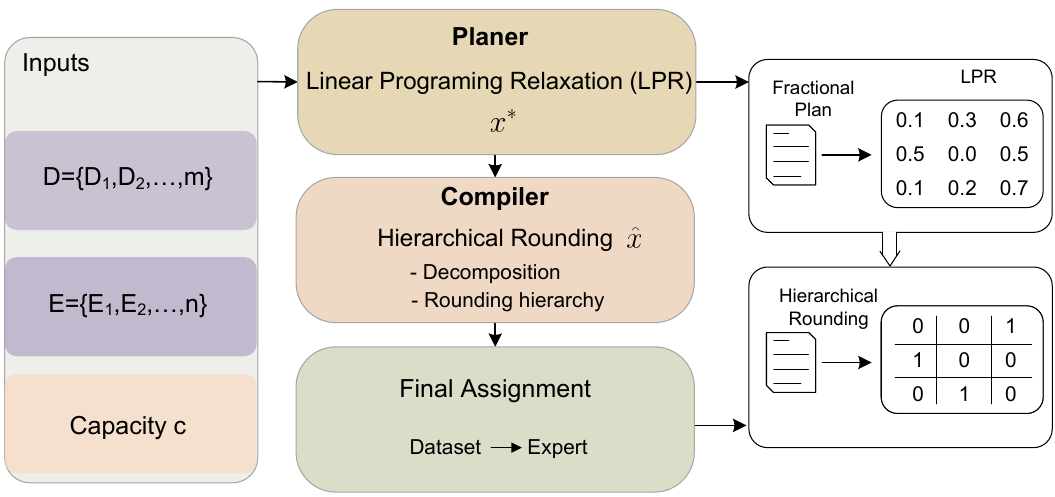}
    \caption{Overview of GEM routing.  Given datasets D, experts E, and expert capacities c,  GEM formulates dataset-expert assignment as a two-stage planner-compiler process. (1) The planner solves a linear program (LP) to produce a fractional assignment $x^\star$; (2) The compiler applies hierarchical (bit-scaling) rounding algorithm, yielding an integral assignment $\hat x$, for dataset-expert mapping. }
    \label{arch}
\end{figure}

Let $D{=}\{D_i\}_{i=1}^m$ be a set of datasets and $E{=}\{E_j\}_{j=1}^n$ experts, where \(m\!>\!n\). For each pair $(D_i, E_j)$ we compute a non-negative affinity score (cost), $w_{ij} \geq 0$, using a frozen backbone (details in Section~\ref{train}), which reflects how well expert $E_j$ is suited to process $D_i$. Our goal is to find a mapping $\rho\!:\!\{D_i\}\to\{E_j\}$ that maximizes total affinity $\sum_{i=1}^m w_{i,\rho(D_i)}$, while respecting specialization factor $c_j\!=\!\lceil m/n \rceil$. This setup aligns with the generalized assignment problem \cite{shmoys1993approximation, bhattacharya2023dynamic, roy2017hierarchical}, which we solve using a two-step planner-compiler strategy, illustrated in Fig.~\ref{arch}. \\

\noindent{\(\blacktriangleright\) \bf{LP Relaxation (Planner):}} The first step computes a fractional solution using linear programming (LP) relaxation. This planner splits a dataset across experts to maximize the total affinity score
%%%%%%%%%%%%%%%%%%%%
\begin{equation} 
\max_{x\in [0,1]^{m\times n}} \sum_{i=1}^m \sum_{j=1}^n w_{ij} x_{ij}. 
\label{LP}
\end{equation}
%%%%%%%%%%%%%%%%%%%%
This satisfies two rules: every dataset is fully assigned,  $\sum_{j=1}^n x_{ij} {=} 1$ for $i\!=\!1, \ldots, m$, and no expert capacity is exceeded, $\sum_{i=1}^m x_{ij}\!\leq\!c_j$ for $j\!=\! 1, \ldots, n$. Importantly, this planner can make globally strategic trade-offs, for example, assigning a dataset to its second-best expert to preserve capacity on another expert that is critical for a higher-affinity dataset. However, the resulting solution $x^\star$ is a soft plan (e.g., $x^\star_{i1}=0.3$, $x^\star_{i2}=0.7$), not an executable schedule. Because we cannot route only 70\% of $D_i$'s batches to E2 and 30\% to E1, as this blended supervision is exactly what $\mathcal{L}_{\text{balance}}$ forces in standard MoEs, and it would reintroduce the mode collapse. Therefore, a compiler (rounding) is required to convert this blueprint into an executable plan ($\rho$).

\vspace{7pt}

\noindent{\(\blacktriangleright\) \bf{Hierarchical Rounding (Compiler):}} To convert the fractional solution $x^\star$ into a final assignment $\hat{x}\!\in\!\{0,1\}^{m \times n}$, we use a capacity-aware rounding strategy  \cite{bernstein2020deterministic, jambulapati2022regularized}. A naive argmax ($\arg\max_j x^\star_{ij}$) can violate capacity constraints (see Section \ref{assignment}). For example, if expert E1 (with $c\!=\!2$) is the top choice for three datasets, argmax would assign all three, violating $c$. Instead, we propose hierarchical (bit-scaling) rounding algorithm, which decomposes each fractional score $x^\star_{ij}$ into a weighted sum of binary bits and processes them in descending order of significance
%%%%%%%%%%%%%%%%%%%%
\begin{equation}
\begin{aligned}
x^\star_{ij}&=\sum_{k=1}^{L} b^{(k)}_{ij}\,2^{-k},\\
x^\star_{ij}&=\underbrace{b^{(1)}_{ij} \cdot 2^{-1}}_{0.5 \text{ bit}} + \underbrace{b^{(2)}_{ij} \cdot 2^{-2}}_{0.25 \text{ bit}}+ \dots+\underbrace{b^{(L)}_{ij} \cdot 2^{-L}}_{\text{smallest bit}}.
\label{round}
\end{aligned}
\end{equation}
%%%%%%%%%%%%%%%%%%%%
where $b_{ij}^{(k)}\in\{0,1\}$ and $L$ controls the rounding precision. These bits are processed in a coarse-to-fine order, starting from the highest bit ($k{=}1$, weight 0.5). This ensures that high-confidence assignments (those with a fractional score $\ge 0.5$) are considered first. In each round $k$  (see Algorithm \ref{alg:round}), we consider all dataset-expert pairs $(i, j)$ where $b^{(k)}_{ij} = 1$ and $D_i$ is still unassigned; sort them by their affinity $w_{ij}$; and assign the best-matching pairs while respecting the expert capacity $c$. Datasets that are not assigned in a given round defer to later rounds with smaller bit contributions.
This solves the argmax failure. For example, if $x^\star_{ij} = 0.7$, it contributes to multiple rounds (e.g., for $k=1,2,3$). If three datasets all want E1 in the $k=1$ round, and E1 has capacity\  2, the two with highest $w_{ij}$ are selected, and the third defers to a later round (e.g., $k\!=\!2$).  This bit-level decomposition reframes the fractional LP into $L$ binary subproblems, ensuring that, every dataset is assigned ($\sum \hat{x}_{ij}\!=\!1$) and no expert is overloaded ($\sum \hat{x}_{ij} \le c_j$).  Importantly, the resulting plan is a high-quality approximation of the LP solution \cite{lee2014path, bhattacharya2023dynamic}, and in the following proposition \ref{prop} we prove the performance gap between the fractional and rounded assignments is minimal $(1-\varepsilon)$. 

Additionally, this compiler is efficient, running in $O(mn \log(1/\varepsilon))$ time, and scalable: adding a new dataset $D_{m+1}$ only requires re-running the rounding, without retraining the entire model. In summary, our two-stage method, using LP for global planning and hierarchical rounding as a deterministic compiler, produces an interpretable dataset-to-expert schedule. 

\begin{algorithm}[t]
\small
\caption{ Global Expert Mapping (GEM)}
\label{GEM}
\begin{algorithmic}[1]
\State \textbf{Input:} $m$ datasets, $n$ experts, utilities $w \in \mathbb{R}^{m \times n}{\geq 0}$, capacities $c \in \mathbb{Z}^n$
\State \textbf{Output:} Final assignment $\hat{x} \in \{0,1\}^{m \times n}$

\Comment{Step 1: Compute soft/fractional plan}

\State Solve LP (Eq.~\ref{LP})  to get $x^\star$ 
\State Set $L \gets \lceil \log_2 m \rceil$  \Comment{Number of bit levels for precision}
\State Initialize $\hat{x} \gets \mathbf{0}^{m \times n}$, $c' \gets c$, assigned[$i$] $\gets$ false for $i=1,\dots,m$ 

\Comment{Step 2: Decompose each fraction into binary bits}

\For{each $i=1$ to $m$, $j=1$ to $n$} 
    \State Set $b^{(k)}_{ij} \gets \lfloor (x^\star_{ij} \cdot 2^L) / 2^{L-k} \rfloor \mod 2$, $k=1,\dots,L$
\EndFor

\Comment{Step 3: Process bits from coarse (large weight) to fine (small weight)}

\For{each level $k=1$ to $L$} \Comment{Assign datasets using biggest chunks first} 
    \State $P_k \gets \{(w_{ij}, i, j) \mid b^{(k)}_{ij} = 1\}$, sort by $w_{ij}$ descending
    \For{$(w_{ij}, i, j)$ in $P_k$}
        \If{not assigned[$i$] and $c'_j \geq 1$} \Comment{Check capacity}
            \State $\hat{x}_{ij} \gets 1$, assigned[$i$] $\gets$ true, $c'_j \gets c'_j - 1$
        \EndIf
    \EndFor
\EndFor
\State \Return $\hat{x}$ \Comment{One expert per dataset}
\end{algorithmic}
\end{algorithm}

\vspace{5pt}

\noindent{\(\blacktriangleright\)\textbf{Proposition (Bounded loss under rounding).}}\label{prop}
Let $x^\star$ be the fractional assignment from the LP (with value $OPT$), and $\hat{x}$ the final integral assignment from our rounding algorithm (with value $\widehat{OPT}$). For any precision $\varepsilon > 0$, our algorithm guarantees
%%%%%%%%%%%%%%%%%%%%
\begin{equation}
\widehat{OPT} = \sum_{i,j} w_{ij}\hat{x}_{ij} \;\ge\; (1-\varepsilon) \cdot OPT
\end{equation}
%%%%%%%%%%%%%%%%%%%%
Thus, the compiled plan preserves at least a $(1-\varepsilon)$ fraction of the LP solution, ensuring minimal degradation during rounding. 

\noindent\textbf{Proof.} First, we define a truncated plan $x^{(L)}$ by keeping $L$ bits of the fractional plan $x^\star$. We then show that our rounding algorithm $\hat{x}$ constructs a final plan that is as good as $x^{(L)}$. A fractional $x_{ij}^\star$ can be written as the series $x_{ij}^\star=\sum_{k=1}^\infty b_{ij}^{(k)}2^{-k}$, thus, our truncated plan is $x^{(L)}_{ij}=\sum_{k=1}^L b_{ij}^{(k)}2^{-k}$ (we choose $L$ based only on inputs ($m, n$) by \(L=\lceil \log_2 (mn / \varepsilon) \rceil\)). The error for each individual assignment $x^\star_{ij}-x^{(L)}_{ij}$ is bounded by $2^{-L}$. Since $w_{ij} \le 1$, the total lost by this truncation is
%%%%%%%%%%%%%%%%%%%%
\begin{equation}
OPT - \sum_{i,j} w_{ij} x^{(L)}_{ij} \le \sum_{i=1}^m \sum_{j=1}^n (1) \cdot 2^{-L} = mn \cdot 2^{-L}
\end{equation}
%%%%%%%%%%%%%%%%%%%%
By setting the $L$, this total truncation error $mn \cdot 2^{-L}$ is $\le \varepsilon$. Assuming $OPT \ge 1$ (which is reasonable, as $OPT \ge \max(w_{ij})$), we have $\varepsilon \le \varepsilon \cdot OPT$. This gives us our first inequality 
%%%%%%%%%%%%%%%%%%%%
\begin{equation}
\sum_{i,j} w_{ij} x_{ij}^{(L)} \ge OPT - \varepsilon \cdot OPT.
\label{partA}
\end{equation}
%%%%%%%%%%%%%%%%%%%%
We now show our plan $\hat{x}$ is as good as the truncated plan $x^{(L)}$
%%%%%%%%%%%%%%%%%%%%
\begin{equation}
\widehat{OPT} = \sum_{i,j} w_{ij} \hat{x}_{ij} \ge \sum_{i,j} w_{ij} x^{(L)}_{ij}.
\label{partB}
\end{equation}
%%%%%%%%%%%%%%%%%%%%
This holds because the algorithm iterates from the most significant bit \((k=1)\) to the least \((k=L)\). At each step, it assigns the candidates with the highest affinity ($w{ij}$) first, while respecting capacity. Any utility from the $x^{(L)}$ plan is only lost if that expert's capacity was filled by an even higher-affinity assignment in the same or a previous round. Therefore, the algorithm preserves the highest-value components of the truncated plan \cite{jambulapati2022regularized, chen2025entropy}. By combining the results from Eqs. \ref{partA} and \ref{partB}, we have $\widehat{OPT} \ge \sum_{i,j} w_{ij} x^{(L)}_{ij} \ge OPT - \varepsilon \cdot OPT = (1-\varepsilon) \cdot OPT$. The hierarchical rounding guarantees that the final assignment achieves a $(1-\varepsilon)$ fraction of the LP, that MoE routing cannot provide.

%%%%%%%%%%%%%%%%%%%%%%%%%%%%%%%%%%%   Algorithm-1  %%%%%%%%%%%%%%%%%%%%%%%%

%%%%%%%%%%%%%%%%%%%%%%%%%%%%%%%%%%%%%%%%%%%%%%%%%%%%%%%%%%%%

\vspace{-5pt}

\subsection{Computation Complexity}
The GEM planner-compiler models routing as a mapping, $\rho: \{1,\dots,m\}\to\{1,\dots,n\}$. The planner solves a linear program with $mn$ variables to find the fractional plan. This step is efficient using modern solvers based on the dual simplex method \cite{jambulapati2022regularized, roy2017hierarchical}. The compiler (hierarchical rounding) then converts this plan into the integral assignment in $O(mn\log(1/\varepsilon))$ time, where $\varepsilon$ is the rounding precision \cite{jambulapati2022regularized,fischer2020improved}. For typical $m$ and $n$, both planning and compiling are negligible setup costs. During inference, GEM's routing reduces to a deterministic $O(1)$ dictionary lookup, activating one expert/GPU path. In each batch, a standard MoE must compute the gating network (involving matrix multiplications, $O(bdn)$ for batch size $b$, feature dimension $d$, and $n$ experts) \cite{shazeer2017sparsely, li2023sparse}; evaluate the load-balancing term, which adding further computational steps and instability \cite{hayes2024buffer, oldfield2024multilinear};  and perform all-to-all token scattering across GPUs \cite{fedus2022switch}. GEM eliminates these bottlenecks by replacing ($G(x) + \mathcal{L}_{\text{balance}}$) with its map $\rho$. %This design activates only one expert (and GPU path) per dataset.  %This non-differentiable map serving as a stable guide for the (differentiable) training of the DINO model \cite{zhang2023dino}. 

\begin{table*}
\centering
\begin{minipage}{0.47\textwidth} 
%\label{data}
\begin{tabular}{@{}c l|ccl@{}} 
\toprule
Index&Data & class & train/test & domain \\ \midrule
1&COCO  \cite{lin2014microsoft}& 80& 35k/5k &natural \\ 
2&LVIS   \cite{gupta2019lvis} &1k & 100k/20k & general objects \\
3&VOC   \cite{everingham2015pascal}& 20& 16k/5k& general objects \\ 
4&WiderFace \cite{yang2016wider}& 60 & 16k/16k& face \\ 
5&DeepLesion  \cite{yan2018deep}& 8& 28k/5k& medical \\
6&Kitchen \cite{kitchen-object-detection-acyvk_dataset}& 14& 5k/2k& indoor \\ 
7&Comic \cite{inoue2018cross}& 6& 1k/1k& comic-book \\ 
8&ClipArt   \cite{inoue2018cross}& 20& 0.5k/0.5k& clipart \\ 
9&DOTA   \cite{xia2018dota}& 15& 19k/10k& aerial \\ 
10&Watercolor  \cite{inoue2018cross}& 6& 1k/1k& watercolor \\ 
11&LISA   \cite{mogelmose2012vision}& 47& 8k/2k& traffic \\ 
12&KITTI  \cite{geiger2012we} &9& 7k/7k& traffic\\ 
\bottomrule
    \end{tabular}
 %  \caption{Universal Object Detection Benchmark (UODB) \cite{wang2019towards}. }
       \end{minipage}  
  % \hfill
 \begin{minipage}{0.4\textwidth}
        \centering
        \includegraphics[width=0.9\columnwidth]{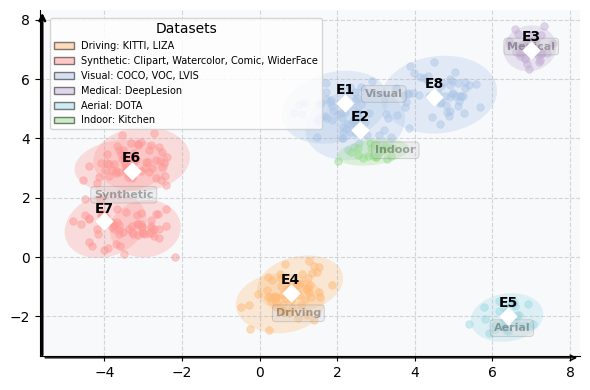}
    \end{minipage}
   \caption{Multi-dataset benchmark from visually and semantically distinct domains, with significant scale imbalance in class counts and dataset sizes. The visualization is isolated domain clusters, showing that a single expert cannot cover all domains. }
  \label{dataset}
\end{table*}

\vspace{-5pt}

\subsection{Architecture}\label{train}
We integrate the GEM planner into the DINO object detector \cite{zhang2023dino} with a standard configuration (ResNet-50 backbone, 6-layer encoder/decoder, hidden dim $=256$). Training runs on 8 RTX-3090 GPUs using a one-expert-per-GPU design. GEM produces a dataset$\!\to\!$expert map via  affinity scores (cost) $w_{ij}$. We pass a small calibration split through a frozen backbone; compute per-dataset mean features, and cluster the pooled features with k-means++ \cite{arthur2007kmeans} to obtain $n=8$ expert anchors. The cost $w_{ij}$ is the cosine similarity between a dataset's mean feature and an expert's anchor. We then solve the LPR and apply hierarchical rounding ($\varepsilon=0.01$) to find the final map $\rho$, respecting the task scheduler $c_j=m/n$. This map is then used to enforce specialization in two variants:

\emph{Variant 1: GEM-DINO (Deep expert specialization).}
In this design, we replace the standard FFNs in the DINO decoder with a GEM module containing $n$ parallel experts. For a batch from dataset $D_i$, the active expert is $j=\rho(D_i)$. Inside each GEM layer, all decoder-query features are routed only to expert $E_j$ (which resides on its own $GPU_j$). The activated parameters per forward pass are thus comparable to the original DINO. This architecture eliminates all routing overhead (gating computation and balancing loss), making it faster than standard MoE implementations.

\emph{Variant 2: GEMGR-DINO (Guided query specialization).}
This variant keeps the DINO encoder and decoder layers unchanged. Specialization is applied at the input (queries) and output (prediction heads) of the decoder. We define $n\!=\!8$ distinct sets of object queries $(Q_1, \dots, Q_n)$, and $n$ prediction heads $(H_1, \dots, H_n)$. The GEM plan $\rho$ guides the entire pass: for dataset $D_i$ with $j=\rho(D_i)$, we feed only $Q_j$ into the decoder, and the decoder's output features are passed only to the corresponding head $H_j$ to generate detections. FLOPs are comparable to DINO ($<$5\% difference), with a moderate parameter increase (46.5M$\rightarrow$49.7M) due to replicated queries/heads.  Both variants use total batch size $16$ and a weighted dataloader sampling $D_i$ with probability $1/\sqrt{|D_i|}$. We train with AdamW (weight decay $0.05$), learning rates $1 \times 10^{-4}$ for transformers and $1 \times 10^{-5}$ for backbone. No auxiliary router or load-balancing losses are required. 

Table \ref{dataset}-(right) shows feature projections with datasets (dots) and experts (E1-E8). Fig. \ref{mapping} shows LP’s fractional (blue) and the final assignments after hierarchical rounding with capacity $c_j$.  

%%%%%%%%%%%%%%%%%%%%%%%%%%%%%%%%%%%%  Algorithm-2  %%%%%%%%%%%%%%%%%%

\begin{algorithm}[t]\small
\caption{Hierarchical Rounding (Compiler)}
\label{alg:round}
\begin{algorithmic}[1]
\State \textbf{Input:} Fractional assignment $x^\star$, expert capacities $C$, bit precision $L$
\State \textbf{Output:} Integral assignment $\hat{x} \in \{0,1\}^{m \times n}$

\State Initialize $\hat{x} \gets 0$, capacity tracker $cap \gets C$, assignment flags $assigned \gets \text{False}$

\For{$k = 1$ to $L$}  \Comment{Process bits from most to least}
    \State Build candidate list of $(i, j)$ where $k$-th bit of $x^\star_{ij}$ is 1 and \\ $assigned[i] = \text{False}$
    \State Sort candidates by $w_{ij}$ in descending order
    \For{$t = 1$ to length of candidate list}
        \State $(i, j) \gets$ candidate at index $t$
        \If{$assigned[i] = \text{False}$ and $cap[j] > 0$}
            \State $\hat{x}_{ij} \gets 1$; $assigned[i] \gets \text{True}$; \\  $cap[j] \gets cap[j] - 1$
        \EndIf
    \EndFor
\EndFor

\For{$i = 1$ to $m$}
    \If{$assigned[i] = \text{False}$}
        \State Assign $i$ to best available $j$ by $w_{ij}$
    \EndIf
\EndFor

\State \Return $\hat{x}$
\end{algorithmic}
\end{algorithm}

%%%%%%%%%%%%%%%%%%%%%%%%%%%%%%%%%%%%%%%%%%%%%%%%%%%%%%%%%%%%

\section{Experiments} %\vspace{-3pt}
\subsection{Datasets and Evaluation Protocol}
\begin{figure}
\centering
   \includegraphics[width=0.97\columnwidth]{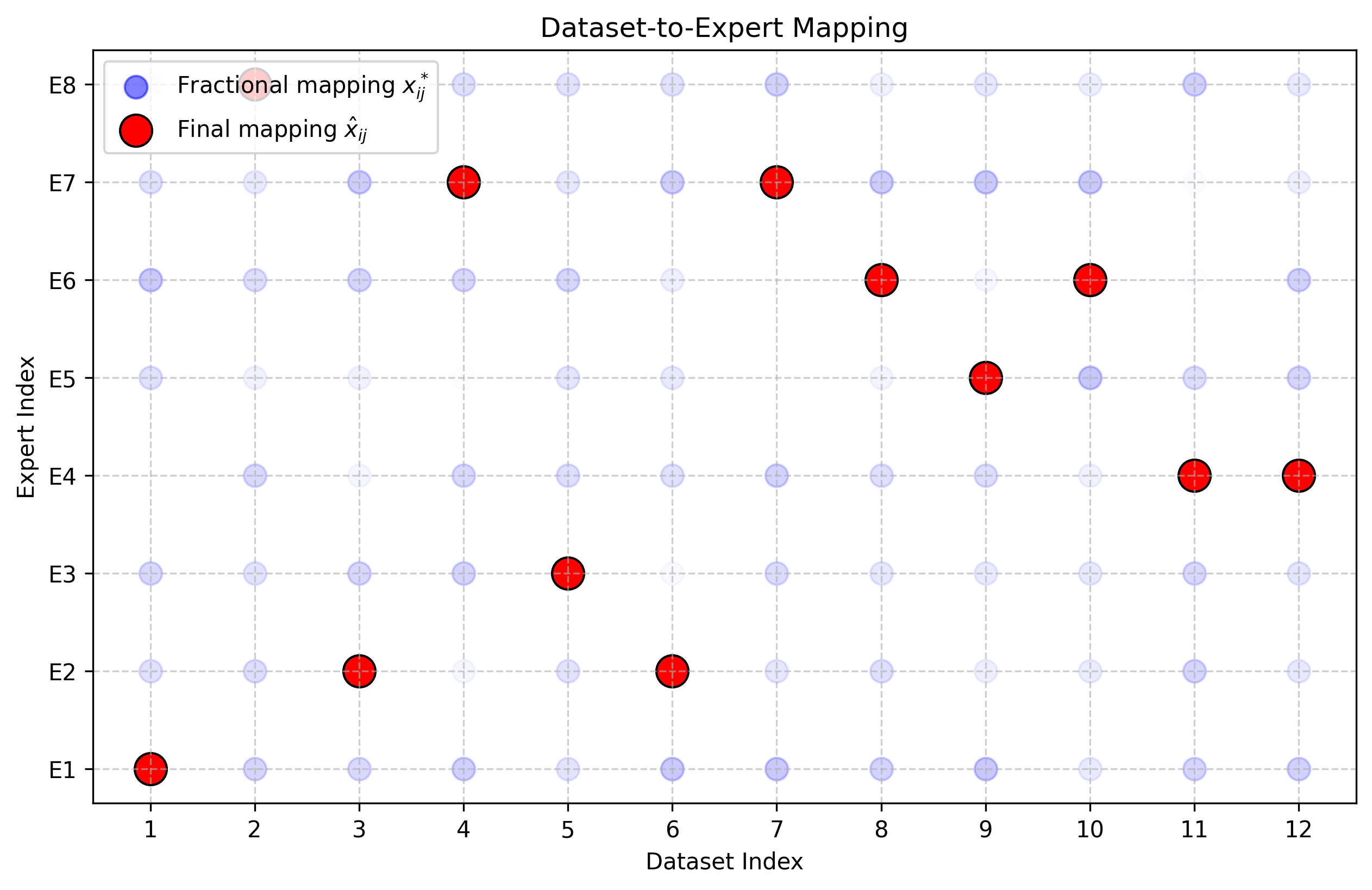}
 \caption{Dataset-Expert mapping under GEM for 12 datasets (IDs in Table~\ref{dataset}) across 8 experts. Blue circles denote the fractional plan $x^\star$, where datasets may split across multiple experts. Red markers show the final assignment $\hat{x}$ after hierarchical rounding. }
   \label{mapping}
\end{figure}

We evaluate GEM-DINO on the Universal Object Detection Benchmark (UODB) \cite{wang2019towards}, which comprising 11 diverse datasets: COCO, VOC, KITTI, LISA, DOTA, DeepLesion, Kitchen, WiderFace, Comic, Clipart, and Watercolor (see Table \ref{dataset}). To further test scalability, we also add the LVIS v1 \cite{gupta2019lvis} (training set; $\sim$100k images, $\sim$1k categories), yielding a mixture of  $m=12$ datasets.  We report the mean Average Precision (AP) on the validation or test set of each corresponding dataset.  With \(m{=}12\) datasets, \(n{=}8\) experts, and capacity \(c_j\!=\!m/n\), our model produces the following  mapping: 
\(
E_1:\text{COCO};\ 
E_2:\text{VOC, Kitchen};\ 
E_3:\text{DeepLesion};\ 
E_4:\text{KITTI, LISA};\ 
E_5:\text{DOTA};\ 
E_6:\text{Clipart, Watercolor};\ 
E_7:\text{Comic, WIDERFace};\ 
E_8:\text{LVIS}.
\)
Fig. \ref{mapping} visualizes this result. GEM acts as an optimal balancer, grouping high-affinity tasks (up to $c_j{=}2$ datasets per expert in this setting) and ensuring all 8 GPUs are fully utilized.
%%%%%%%%%%%%%%%%%%%%
%%%%%%%%%%%%%%%%%%%%
We conduct three main experimental evaluations: (i) multi-domain benchmarking to compare GEM-DINO with strong baselines and analyzing expert specialization; (ii) applying GEM to fine-grained category mapping (e.g., 500+ categories) across a fixed number of experts $n\!=\!8$, to assess the robustness of the planner under high granularity; and (iii) replacing softmax-based gating in Vision-MoE \cite{riquelme2021scaling} with GEM’s deterministic plan, and comparing against SOTA routing methods.

%%%%%%%%%%%%%%%%%%%%%%%%%%%%%%%%%%%%%%%%%%%%%%%%%%%%%%%%%%%%%%%%%%

\subsection{Baselines} 
We compare our model against three baselines, all built on the same DINO (ResNet-50) architecture. The single-dataset baseline trained separately on each individual dataset. The mixed-dataset is a standard DINO trained on the combined mixture of all datasets, representing the non-MoE, joint-training approach. Finally, the MoE-DINO replaces the decoder FFN layers with standard MoE \cite{puigcerver2023sparse} layers and is trained with the load-balancing loss ($\mathcal{L}_{\text{balance}}$).

%%%%%%%%%%%%%%%%%%%%%%%%%%%%%%%%%%%%%%%%%%%%%%%%%%%%%%%%%%%%%%%%

\setlength{\tabcolsep}{3pt}
\begin{table*}
\centering
\begin{tabular}{l|c|cccccccccccc|c} \toprule
Method & Param. & LVIS& KITTI &VOC &WiderFace& LISA& Kitchen &COCO & DOTA &DLesion &Comic& ClipArt& WaterColor& Avg.\\  \midrule
Shi et al. \cite{shi2024plain} & 43.5M & 27.2 & 61.5 & 53.7 & 35.4 & 68.2 & 46.5 & 38.3 & 41.5 & 38.3 & 24.8 & 24.2 & 27.5 & \color{gray}40.6 \\
Zhou et al. \cite{zhou2022simple}& 69.3M & - &49.2 & 31.8& 27.5 & 61.8& 43.1& 19.7& 29.5& 23.2 & 24.1& 23.9& 26.3& \color{gray}32.7  \\ \midrule
\rowcolor{orange!10}
Single-DINO & 46.5M & 31.3 & 69.0 & 57.8 & 39.3 & 76.9 & 47.3 & 42.6 &44.1 &40.5 &13.9 & 11.7 & 16.3 & \color{gray} 40.8 \\
\midrule
Mixed-DINO \cite{zhang2023dino} & 46.8M & 26.9 & 63.7 & 55.6 & 37.5 & 71.5 & 48.2 & 39.0 & 43.7 & 42.3 & 24.2 & 23.5 & 27.0 & \color{gray} 41.9 \\
DINO-MoE & 47.5M & 27.5 & 64.3 & 57.2 & {38.6} & 72.3 & 50.2 & 39.5 & 44.3 & 42.6 & 25.7& 24.2 & 26.5 & \color{gray}42.7  \\
G-DINO \cite{liu2024grounding} & 56.9M  & \bf{30.2} & 64.6 & {57.9} & 38.5 & 71.8 & 49.7 & {40.6} & \underline{45.8} & 42.4 & {25.9} & 25.1 & 28.3 & \color{gray}43.4  \\ 
DAMEX \cite{jain2024damex} & 47.5M & 27.2 & {65.1} & {56.8} &  {39.0} & 72.6 & 49.4 & {40.2} & {44.9} & {43.0} & {26.1} &{24.8} & 27.6 & \color{gray}43.2  \\ 
Sinkhorn-MoE \cite{clark2022unified} & 48.3M & 27.6 & 65.4 & 57.0& \underline{40.2} & 73.1& 50.2 & {39.7} & {45.0} & {42.7}& \underline{26.3} & \underline{26.0} & 29.3 & \color{gray}43.5   \\
MoE++ \cite{clark2022unified} & 48.1M & {27.9} & \underline{66.5} & {57.4} & {39.2} & \underline{73.6} & 50.8 & {40.5} & 45.7 & \underline{43.4} & {25.8} & {25.2} & 28.0 &  \color{gray}43.6 \\ 
REMoE \cite{wang2024remoe}& 47.6M & {28.3} & {66.2} & \underline{57.8} & {39.5} & {73.4} & \bf51.7 & \underline{41.0} &  {45.6} & {42.9} & \underline{26.3} & {25.7} & 29.2 &  \color{gray}44.0 \\
$\mu$MoE \cite{oldfield2024multilinear} & 47.0M & 28.5 & 65.9 & 57.3 & 38.7 & {72.9} & 49.5 & {40.3} & 44.8 & {42.5} & 25.7 & {25.2} & 28.6 &\color{gray} 43.3 \\ \midrule 
\rowcolor{teal!5} 
GEMGR-DINO & 49.7M & 28.4 &66.5& 57.9 & 40.1& 73.4& 51.0& 40.6 & 45.7& 43.0& 26.2 & 26.4 & \bf31.5 & \color{gray} 44.2 \\ \rowcolor{teal!15} 
GEM-DINO & 47.3M & \underline{29.2} & \bf67.1 & \bf 58.1 & \bf 40.4 & \bf74.2 & \underline{51.4} & \bf 41.5 & \bf 46.0 & \bf43.8 &\bf 27.5 & \bf{26.8} & \underline{30.9} & \color{gray} 44.8 \\
\bottomrule
    \end{tabular}
    \caption{Performance of the UODB. We report the mean AP across 12  datasets.  Single-DINO is added as a reference, which means trained on each dataset separately. Bold indicate the best performance per dataset, underline the second-best. }
    \label{tab1}
\end{table*}
%%%%%%%%%%%%%%%%%%%%%%%%%%%%%%%%%%%%%%%%%%%%%%%%%%%%%%%%%
\begin{figure*}
\centering
\includegraphics[width=2.02\columnwidth]{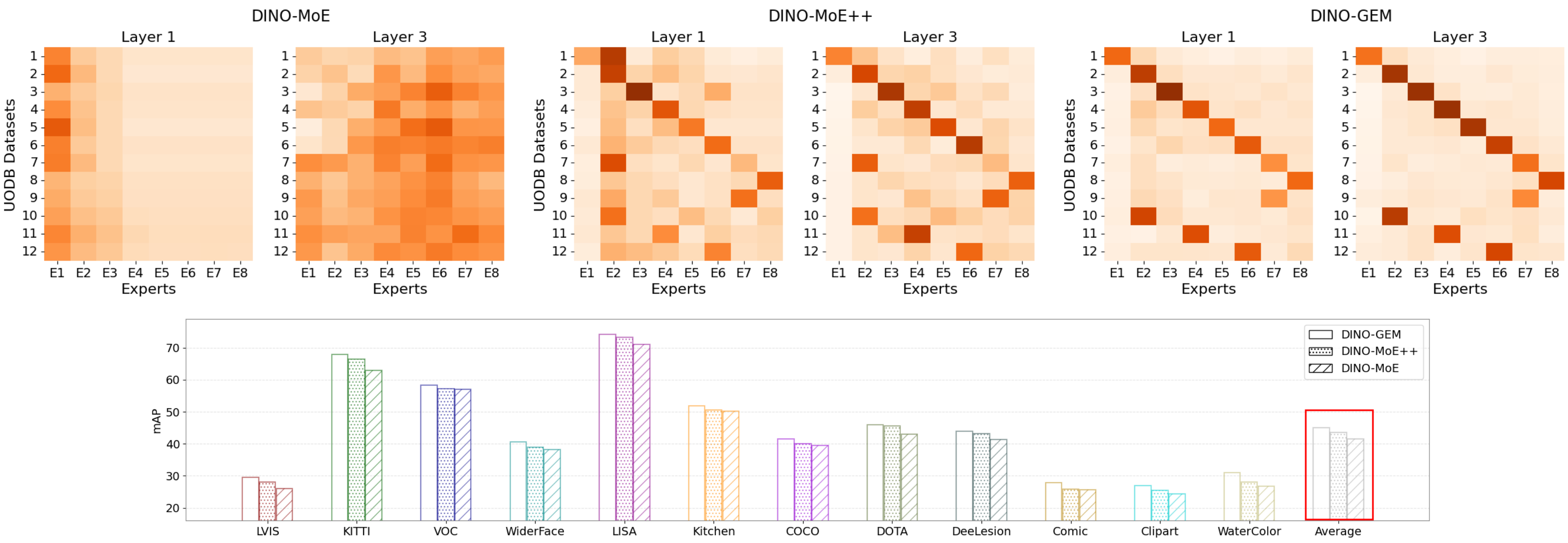} 
\caption{Visualization of expert routing and its effect on performance. (Top) Heatmaps showing dataset-to-expert routing at decoder layers. Rows correspond to datasets, columns to experts. DINO-MoE shows diffuse and overlapping routing, indicating mode collapse caused by its balancing loss (i.e., all datasets are routed to all experts). DINO-MoE++ mitigates collapse via capacity constraints but still results in blurry, less interpretable routing. GEM-DINO produces sharp, stable, and interpretable plan, even in early layers. (Bottom) Per-dataset AP comparison shows that GEM-DINO achieves the best overall performance, particularly on underrepresented domains. }  
\label{user}
\end{figure*}
%%%%%%%%%%%%%%%%%%%%%%%%%%%%%%%%%%%%%%%%%%%%%%%%%%%%%%%%

\vspace{-3pt}

\subsection{Multi-domain Detection Performance}
We evaluate our models on 12 diverse datasets (see Table~\ref{dataset}) and report results in Table~\ref{tab1}. We compare with general multi-dataset detectors \cite{shi2024plain, zhou2022simple, liu2024grounding}, single-DINO, mixed-DINO, and state-of-the-art MoE routers (SoftMoE \cite{puigcerver2023sparse}, Sinkhorn-MoE \cite{clark2022unified}, MoE++ \cite{jin2024moe++}, REMoE \cite{wang2024remoe}, DAMEX~\cite{jain2024damex}, and $\mu$MoE~\cite{oldfield2024multilinear}). For fairness, all MoE variants are implemented within the DINO. In particular, REMoE replaces the top-k gating in the standard MoE with a ReLU-based router \cite{wang2024remoe} and is trained with L1 regularization (instead of $\mathcal{L}_{\text{balance}}$ loss) to encourage sparse expert selection. This setup represents a recent effort to address expert collapse without uniform balancing. The results illustrate the core challenge of multi-domain learning. 
The single-DINO (trained per-dataset) performs well on large datasets by avoiding cross-domain interference but fails on small domains due to limited training data. The mixed-DINO (joint training) solves the rare-domain problem through joint training but degrading performance on large datasets (39.0 on COCO vs. 42.6 for single-DINO). The MoE routers attempt to solve this trade-off. Sinkhorn-MoE uses optimal transport to enforce uniform token-to-expert routing, improving results on small datasets (e.g., Comic), but diluting specialization on more complex domains like DOTA or DeepLesion. MoE++ introduces residual gated FFNs. While the residuals help to learn shared patterns (e.g., KITTI and LISA), it still underperforms on small datasets due to insufficient expert focus.
Our model solves both sides of this conflict. We route low-resource datasets to dedicated experts, protected by capacity constraints (based on mapping $\rho$), avoiding interference from large-scale datasets. This leads to state-of-the-art results on these domains where all other methods struggle. Fig. \ref{user} visualizes this deterministic plan, showing stable and balanced expert usage across all datasets. 
GEM also demonstrates superior parameter efficiency. It achieves higher accuracy than Grounding-DINO (with a Swin-T backbone), while requiring about 30\% fewer parameters. This suggests our routing is a more parameter-efficient alternative to scaling up the backbone. We compute the AP-per-parameter ratio for the top models, and GEM-DINO ($44.8/47.3\!\approx\!0.95$), is higher than REMoE ($44.0/47.8\!\approx\!0.92$) or MoE++ ($43.7/48.1\!\approx\!0.91$), confirming it achieves superior accuracy at a lower parameter cost. While GEMGR variant also performs well (44.2 AP), we focus our analysis on GEM-DINO  due to its superior parameter efficiency.

%%%%%%%%%%%%%%%%%%%%%%%%%%%%%%%%%%%%%%%%%%%%%%%%%%%%%%%%%%%%%%
\begin{figure*}[t]
    \centering
    \includegraphics[width=0.887\linewidth]{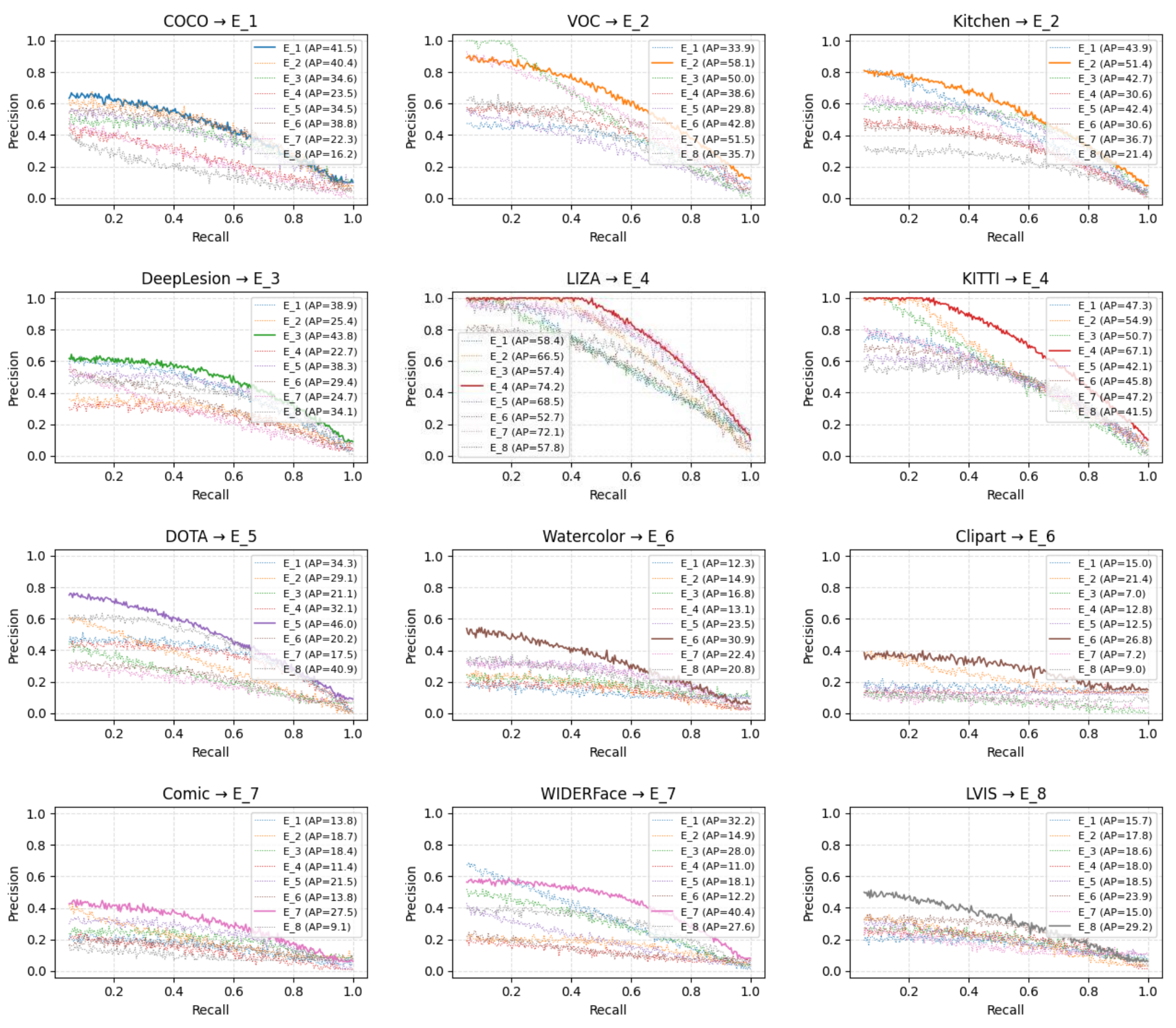}
    \caption{Expert specialization analysis of our GEM-DINO. For each of the 12 datasets (UODB+LVIS), we evaluate detection performance (AP vs. recall) using all 8 expert individually, $E_1,\ldots,E_8$. Each plot corresponds to one dataset. The expert assigned by GEM (via LP + hierarchical rounding) is shown in solid, while all other non-assigned experts are dashed. Across datasets, the assigned expert consistently outperforms others, validating that our planner correctly identified the optimal dataset-to-expert mapping. }
    \label{PR}
\end{figure*}
%%%%%%%%%%%%%%%%%%%%%%%%%%%%%%%%%%%%%%%%%%%%%%%%%%%%%%%%%%%%%%
\begin{table}[t]
\centering
\setlength{\tabcolsep}{5pt}
\begin{tabular}{l|cc}
\toprule
Method & COCO & DOTA \\
\midrule
DINO         & 41.6 & 43.7 \\
SoftMoE     & 41.8 (\textcolor{ForestGreen}{+0.2 / 0.5\%}) & 44.0 (\textcolor{ForestGreen}{+0.3 / 0.7\%}) \\
Sinkhorn-MoE & 41.9 (\textcolor{ForestGreen}{+0.3 / 0.7\%}) & 44.5 (\textcolor{ForestGreen}{+0.8 / 1.8\%}) \\
REMoE        & 42.3 (\textcolor{ForestGreen}{+0.7 / 1.8\%}) & 44.9 (\textcolor{ForestGreen}{+1.2 / 2.7\%}) \\
\rowcolor{teal!10}
GEM     & {42.5} (\textcolor{ForestGreen}{{+0.9 / 2.2\%}}) &
               {45.3} (\textcolor{ForestGreen}{{+1.6 / 3.7\%}}) \\
\bottomrule
\end{tabular}
\caption{Domain-shift case: COCO and DOTA pose a strong domain shift (ground-level vs. aerial views). MoE variants improve over DINO, and GEM-DINO achieves the largest gain. } \label{domain1}
\end{table}
\begin{table}
\centering
%\small
\setlength{\tabcolsep}{5pt}
\begin{tabular}{l|cc}
\toprule
Method & LISA & KITTI \\
\midrule
DINO         & 74.9 & 67.5 \\
SoftMoE     & 75.6 (\textcolor{ForestGreen}{+0.7 / 0.9\%}) & 68.1 (\textcolor{ForestGreen}{+0.6 / 0.9\%}) \\
Sinkhorn-MoE & 76.0 (\textcolor{ForestGreen}{+1.1 / 1.5\%}) & 68.4 (\textcolor{ForestGreen}{+0.9 / 1.3\%}) \\
REMoE        & 75.7 (\textcolor{ForestGreen}{+0.8 / 1.2\%}) & 69.0 (\textcolor{ForestGreen}{+1.5 / 2.2\%}) \\
\rowcolor{teal!10}
GEM     & {76.2} (\textcolor{ForestGreen}{{+1.3 / 1.8\%}}) &
               {69.2} (\textcolor{ForestGreen}{{+1.7 / 2.5\%}}) \\
\bottomrule
\end{tabular}
\caption{Label-space mismatch case: LISA (traffic signs) vs. KITTI (vehicles). MoE routing reduces negative transfer from disjoint labels, and parentheses show improvement vs.\ DINO. } 
\label{domain2}
\end{table}
%%%%%%%%%%%%%%%%%%%%%%%%%%%%%%%%%%%%%%%%%%%%%%%%%%%%%%%%%%%%

\begin{table}
\centering
%\small
\setlength{\tabcolsep}{4pt}
\begin{minipage}{0.6\textwidth}
\begin{tabular}{l   c   c   c   c}
\toprule
Model & 50 shots & 100 shots & 1000 shots  & full  \\
\midrule
DINO          & 11.9\, / \,56.0  & 18.7\, / \,55.8 & 33.9\, / \,56.4 & 37.5\, / \,56.5 \\
DINO-MoE      & 13.6\, / \,55.9 & 20.2\, / \,55.8 & 36.4\, / \,56.7 & 38.2\, / \,57.0 \\
MoE++         & 14.7\, / \,56.4 & 22.8\, / \,56.7 & 37.9\, / \,57.2 & 38.6\, / \,56.9 \\
Sinkhorn-MoE  & 15.2\, / \,56.2 & 23.5\, / \,56.6 & 37.5\, / \,57.0 & 39.4\, / \,57.3 \\\rowcolor{teal!10}
\textbf{GEM-DINO} & 16.4\, / \,57.0
                  & 24.5\, / \,56.9
                  & 38.2\, / \,57.3
                  & 39.6\, / \,57.5 \\
\bottomrule
%\caption{}
%%%\label{fewshot}
        \end{tabular}
  %     \caption*{}
    %   \label{balanced}
    \end{minipage}
   % \vspace{-10pt}
    \begin{minipage}{0.37\textwidth}
  \centering
    \includegraphics[width=\linewidth]{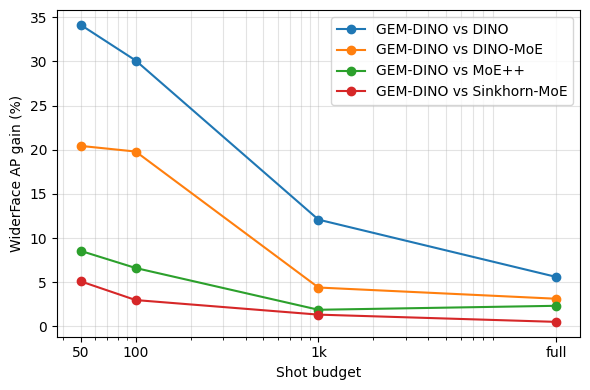}
   %  \caption*{}
     %  \label{aaa}
    \end{minipage}
    \caption{Few-shot adaptation performance on VOC (Base, $\text{AP}_b$) and WiderFace (Target, $\text{AP}_t$), reporting $\text{AP}_t / \text{AP}_b$. The dense DINO struggles in this low-shot regime (low $\text{AP}_t$). MoE variants accelerate target learning but suffer from task interference, leading to a large drop in their base performance $\text{AP}_b$. GEM-DINO achieves the highest target $\text{AP}_t$ while also maintains the base-task stability, with a minimal $\text{AP}_b$ drop, especially at low shot.}
   \label{fewshot}
\end{table}
%%%%%%%%%%%%%%%%%%%%%%%%%%%%%%%%%%%%%%%%%%%%%%%%%%%%%%%%%%%%

\vspace{-3pt}

\subsection{Expert Specialization Analysis}
\begin{figure*}
    \centering
    \includegraphics[width=0.9\linewidth]{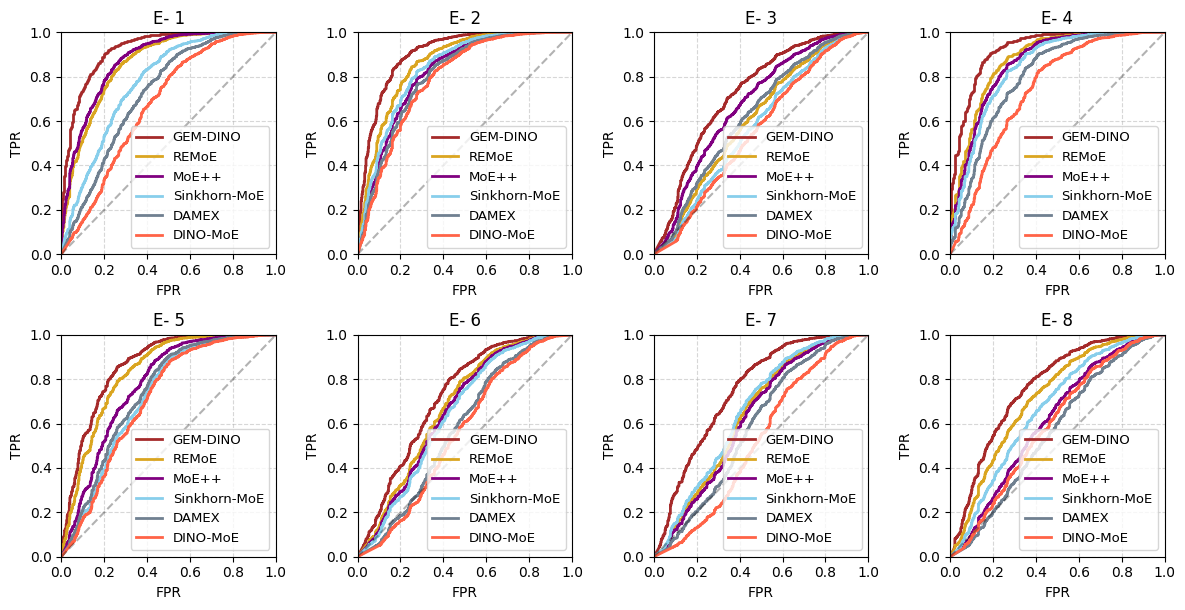}
    \caption{Per-Expert performance comparison. While our main results (Table \ref{tab1}) show a superior average mAP, this plots provide a more granular analysis. As expected, GEM-DINO achieves the best performance (highest TPR at all FPR levels) across all 8 expert slots. It proves that our plan-then-compile strategy produces a more effective specialist for each assigned task than the other MoE routers. }
    \label{ROC}
\end{figure*}

A common challenge with most MoEs is the lack of clear expert specialization \cite{oldfield2024multilinear, omi2025load, hayes2024buffer}. This term refers to the ideal state where each expert becomes a true specialist. For example, in a specialized model, expert 1 would get a high score on medical data, while all other experts (E2-E8) would get a very low score on that same data. This performance gap is the proof of specialization. To validate that GEM solves this, we conduct a cross-expert evaluation in Fig. \ref{PR}. This experiment is designed to answer two questions: (1) Does our GEM planner correctly identify the best expert for each dataset? (2) Does our capacity-aware assignment ($c_j$) co-specialize a single expert on multiple datasets? For this, we train the model on a mixture of 12 datasets and evaluate each dataset independently against all 8 experts. In the plots, the expert assigned by GEM is the solid line (specialist), while all non-assigned experts are dashed lines (non-specialist). In almost every plot, the assigned expert (solid line) achieves the highest AP, demonstrating that our affinity (cost) metric is an accurate predictor of final performance. On challenging domains like DeepLesion (E3) and DOTA (E5), the assigned experts maintain high precision, while the non-assigned experts show a performance drop, with their precision collapsing early. This performance gap proves that the experts are not random or redundant generalists; mode collapse is avoided, and GEM enables interpretable specialization.  The experiment also validates our second question, capacity-aware co-specialization. The planner, guided by $c_j$ constraint, assigned both KITTI and LISA to E4 (both are traffic-scene domains). The plots confirm that E4 is, in fact, the top-performing expert for both datasets. This co-specialization is also seen for E2 (VOC, Kitchen) and E7 (Comic, WIDERFace). 
Fig.~\ref{user} visualizes the internal routing patterns (top) and their impact on performance (bottom). Each heatmap shows the dataset-to-expert assignments for a given model. DINO-MoE displays a diffuse and unfocused pattern, indicating mode collapse, where the balancing loss forces all datasets through all experts. In contrast, GEM-DINO produces a clean, deterministic pattern, confirming our stable assignment of each dataset to a dedicated expert. 
%%%
%
We then quantify the impact of this specialization in Fig. \ref{ROC}. While our main results (Table \ref{tab1}) report an average AP, these plots provide a more granular per-expert analysis against other MoE baselines. A true positive occurs when an expert (e.g., E2) correctly claims an input from a dataset it is assigned to (e.g., VOC or Kitchen). A false positive occurs when E2 incorrectly claims an input from a dataset it is not assigned to (e.g., COCO). Across all eight tasks, GEM-DINO achieves the highest true positive rate for any given false positive rate, demonstrating that our plan-then-compile strategy produces a more effective specialist than other MoE routers.  

\vspace{-5pt}

%%%%%%%%%%%%%%%--------------------------------%%%%%%%%%%%%%%%%%%%%%%%%

\subsection{Few-Shot Adaptation and Task Interference}
A key measure of a multi-domain robustness is its ability to handle data imbalance. We show this by creating a highly imbalanced dataset, using the full VOC dataset (as the base task) paired with $k$-shots (50, 100, 1000, or full) of the WiderFace (as the target task), from the UODB dataset. As shown in Table \ref{fewshot}, we evaluate performance on both the target domain (WiderFace, $AP_t$) and the base domain (VOC, $AP_b$). The dense DINO suffers from task interference. Because the VOC data is dominant, the model's shared weights are optimized almost entirely for VOC.  The rare WiderFace (e.g., 50 shots) are treated as noise, resulting in the worst $AP_t$ (11.9). The MoE routers (MoE, MoE++ and Sinkhorn) try to fix this; as the results show, Sinkhorn-MoE achieves a good $AP_t$ (15.2) but at the cost of sacrificing base performance $AP_b$ (56.2). However, GEM-DINO, at every shot level, achieves the highest $AP_t$ (e.g., 16.4) and the highest $AP_b$ (e.g., 57.0). Indeed, the dense DINO parameters are polluted by gradients from both tasks, while the GEM planner identifies VOC and WiderFace as distinct tasks and assigns them to separate, dedicated expert slots. When a VOC batch is processed, its gradients only update the shared backbone and its own expert. When a WiderFace batch is processed, its gradients only update the shared backbone and dedicated expert. This architecture prevents task interference.

\vspace{-5pt}

\subsection{Domain Shift \ vs.\ Label Mismatch}
We evaluate the primary challenges in multi-dataset learning: domain shift and label mismatch. First, we analyze the more challenging scenario of label-space mismatch (Table \ref{domain2}). We mix LISA and KITTI, which are from a similar traffic domain but have disjoint or missing labels; objects like cars and signs are present in both, but only labeled in one. The dense DINO baseline suffers from negative transfer, as its shared weights receive conflicting gradients; when it trains on a KITTI image, all traffic signs are treated as background, and vice-versa. This forces the model to be confused, degrading performance on both tasks. MoEs provide a partial improvement by routing some data to different experts, but their balancing or soft routing still allows for cross-task interference. However, GEM-DINO's capacity constraint ($c_j$) assigns each dataset to a separate expert. Thus, the LISA expert only sees LISA gradients (and never learns cars are background), while the KITTI expert only sees KITTI gradients (and never learns signs are background). 
Next, we analyze domain shift (Table \ref{domain1}) by mixing datasets with a similar label set but different domains: COCO (ground-level) and DOTA (aerial views). This creates conflicting feature supervision, as objects like vehicle have completely different visual representations. The dense DINO baseline struggles to learn these opposing features in the same set of weights. Again, GEM-DINO provides the largest improvement. By assigning COCO and DOTA to two separate experts, our model eliminates this feature conflict, allowing one expert to specialize on ground-level features and the other on aerial features. In this experiment, all MoEs use the standard DINO as their backbone.

%%%%%%%%%%%%%%%---------------------------------%%%%%%%%%%%%%%%%%%%%

\begin{figure}[h]
    \centering
    \includegraphics[width=0.98\linewidth]{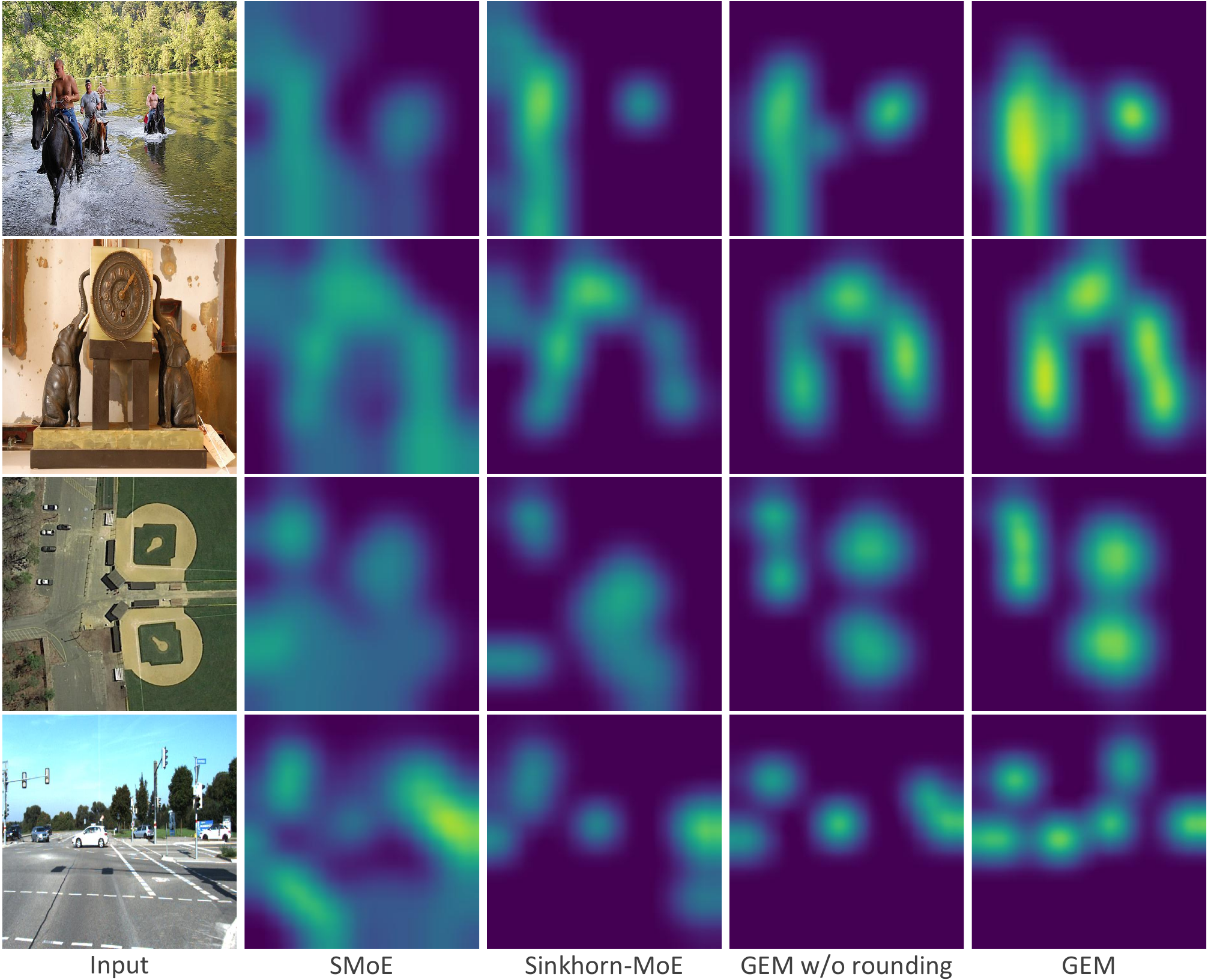}
    \caption{Visualizing expert specialization. Each row shows an input and the corresponding heatmaps from different models. SoftMoE shows diffuse, noisy activations. Sinkhorn yields sharper maps but misses relevant regions. GEM without rounding results in overlapping activations, as fractional assignments distribute each query across multiple experts. GEM produces object-aligned maps, demonstrating accurate and interpretable localization. }
    \label{attention}
\end{figure} 

%%%%%%%%%%%%%%%---------------------------------%%%%%%%%%%%%%%%%%%%%

%%%%%%%%%%%%%%%---------------------------------%%%%%%%%%%%%%%%%%%%%

\setlength{\tabcolsep}{3pt}
\begin{table*}
   \centering
    \begin{tabular}{l|cccccccccccc|c} \toprule
Mapping & LVIS & KITTI &VOC &WiderFace& LISA& Kitchen&COCO & DOTA &DeepLesion &Comic& ClipArt& Watercolor&  Avg.\\  
\midrule
SoftMoE &  27.5 & 64.3 & 57.2 & {38.6} & 72.3 & 50.2 & 39.5 & 44.3 & 42.6 & 25.7& 24.4 & 26.8 & \color{gray}42.7  \\

Random routing  & 27.8 & 64.6 & 57.0 & 38.2 & 71.9 & 49.8 &40.8 & 45.0 & 42.7 & 26.4 & 24.6 & 27.5 &  \color{gray} 42.9 \\ 

W/O Rounding  & 28.6 & 65.8 & 57.3 & 38.5 & 73.2 & 50.3  & 41.2 & 44.9 & 43.0 & 25.8 & 24.7 & 28.2 & \color{gray} 43.5 \\

Fractional compiler  &  28.5 & 66.4 & 57.9 & 39.0 & 74.4 & 51.2 & 40.7 & 45.5 & 42.9 & 26.3 & 25.1 & 27.9 &    \color{gray} 43.9   \\

\rowcolor{teal!10} 

GEM & 29.2 & 67.1 & 58.1 &  40.4 & 74.2 & 51.4 &  41.5 & 46.0 & 43.8 & 27.5 & 26.8 & 30.9 & \color{gray} 44.8 \\
\bottomrule
   \end{tabular}
    \caption{Comparison of routing strategies for expert assignment on multi-dataset object detection benchmark. Random routing outperforms SoftMoE, which includes a load balancing loss; greedy LP assignment (without rounding) and fractional compiler leverage LP scores and yield stronger results, but suffer from expert overload and noisy specialization.}
    \label{map}
\end{table*}

%%%%%%%%%%%%%%%---------------------------------%%%%%%%%%%%%%%%%%%%%

%\vspace{-6pt}

\subsection{Different Mapping Strategies}\label{assignment}
Table \ref{map} compares our GEM framework against several mapping strategies on the UODB benchmark, isolating the contributions of its planner and compiler components.

\vspace{-5pt}

\begin{itemize}[]
    \item SoftMoE (baseline) \cite{puigcerver2023sparse}, uses softmax gating with balancing loss \((\lambda\!=\!0.1)\). While effective in single-domain settings \cite{li2023sparse, krajewski2024scaling}, it struggles in multi-dataset training, hurting performance on low-resource domains.
%%%%%%%%%%%%%%%%%%%%%%%%%%%%%%%%%%%%%%%%
    \item Random mapping with $w_{ij}\!=\!1$, randomly assigns each dataset $D_i$ to expert $E_j$ (assuming uniform affinity scores), while still respecting the $c_j$ constraint. Despite its simplicity, it outperforms SoftMoE, showing the benefit of dataset-level routing (even random) over stochastic gating. 
%%%%%%%%%%%%%%%%%%%%%%%%%%%%%%%%%%%%%%%%%%%%%%%%%%%%%%%
   \item W/O Rounding (argmax), uses only the planner without the compiler. Each dataset $D_i$ simply assigned to its highest-scoring expert ($\hat{\rho}(D_i)\!=\!\arg\max_j x^\star_{ij}$). This fails because it ignores capacity: multiple datasets can all choose the same expert, leading to an overloaded and sub-optimal schedule (see the 4th example in Fig. \ref{attention}).
%%%%%%%%%%%%%%%%%%%%%%%%%%%%%%%%%%%%
    \item Fractional compiler is an expert-centric strategy. Each expert claims the $c_j$ datasets for which it has the highest score $x^\star$. This creates assignment conflicts: a single dataset can be the top choice for multiple experts, leading to an invalid one-to-many map; while a weak dataset might not be in any expert's top $c_j$ list, leaving it unassigned.
%%%%%%%%%%%%%%%%%%%%%%%%%%%%%%%%%%%%%%%%
    \item Our full approach, using the LPR planner and hierarchical rounding, prevents both capacity violations and assignment conflicts, achieving the best overall performance.
    \end{itemize}
Overall, softmax-gating routing struggle to handle multi-domain diversity. Naive routing (random or No-Rounding) leads to collapse and capacity violations, while GEM formulation provides a more effective routing that avoids cross-domain interference.

%%%%%%%%%%%%%%%---------------------------------%%%%%%%%%%%%%%%%%%%%

%\vspace{-2pt}

\subsection{Class-Level Specialization}\label{class-level}
To assess flexibility and scalability, we extend GEM to class-level specialization across hundreds of object categories. This experiment uses a mixture of three large datasets (COCO \cite{lin2014microsoft}, Objects365 (O365) \cite{shao2019objects365}, and Visual Genome (VG) \cite{tang2020unbiased}), where the number of datasets ($m\!=\!3$) is smaller than the number of experts ($n\!=\!8$). We first unify the label spaces into a common class set $\mathcal{C}$ with $|\mathcal{C}|\!\approx\!500$. The goal is to assign classes (not datasets) to experts so that each expert specializes in a coherent subset. We construct an $|\mathcal{C}|\!\times\!n$ affinity matrix $w_{c_j}$ between class $c\!\in\!\mathcal{C}$ and expert $E_j$ and run the GEM planner (LP + hierarchical rounding) to produce a deterministic class$\to$expert map, $\rho: \mathcal{C} \to \{E_1,\ldots,E_8\}$. 
This class-level assignment introduces a new challenge, where a single image may contain multiple objects (e.g., person and car), each mapped to a different expert (e.g., E1 and E2). To handle this, we modify the DINO decoder by replacing its feed-forward layers with a GEM expert module and add a lightweight assignment predictor $A(x)$ (a single linear layer) that outputs a distribution $g_q$ over experts for each decoder query $q$.  This router is not trained with a standard balancing loss. Instead, we use our global plan $\rho$ as the ground-truth supervision. 
During training, if $q$ matches an object of class $c$, the correct expert is $j^\star{=}\rho(c)$ and we optimize $\mathcal{L}_{\text{router}}{=}\text{CrossEntropyLoss}\big(g_q,\ j^\star\big)$. Thus, queries for different classes in the same image can activate different experts, while training remains stable and free of any load-balancing regularizer.

\begin{figure}
   \centering
    \begin{minipage}{0.42\textwidth} 
     % \hspace{-10pt}
        \includegraphics[width=0.9\textwidth]{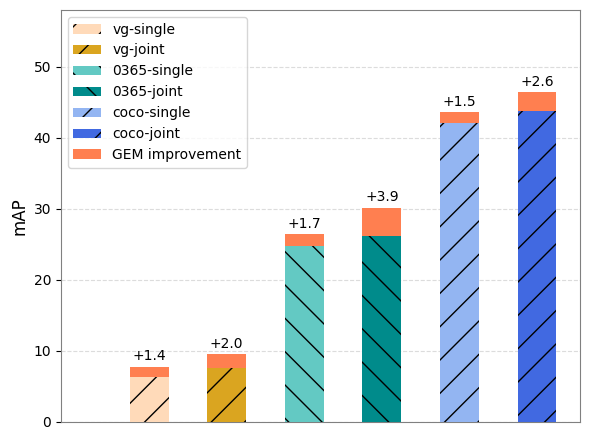} 
    %  \caption{}
    %  \label{nb-data}
    \end{minipage} 
 %  \quad
    \begin{minipage}{0.49\textwidth}
    \centering
    %\hspace{10pt} 
        \includegraphics[width=0.75\textwidth]{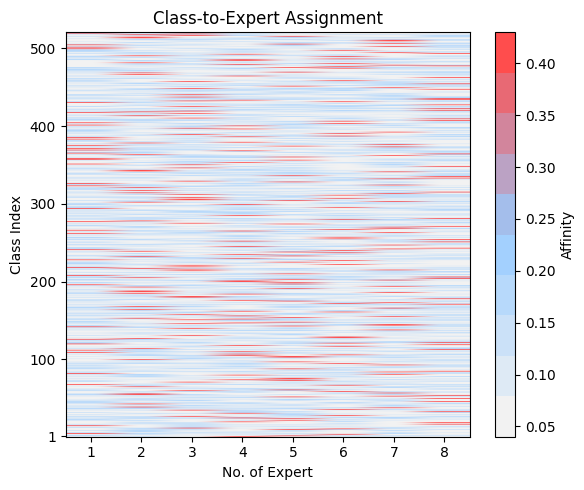}
        \caption{Class-level performance on COCO, O365, and VG. We compare DINO-GEM to baselines trained on either single datasets or a joint mixture. GEM outperforms both, demonstrating fine-grained expert specialization across hundreds of object categories.} 
      \label{expert}
      \end{minipage}
     \end{figure}

Fig. \ref{expert} compares our model with two baselines: single is trained on each dataset individually; joint is trained on the combined datasets. The red stacked bars show the AP improvements of GEM over these baselines. The results show that (1) GEM scales to hundreds of fine-grained tasks (classes); (2) it supports supervised routing in mixed-batch settings without balance losses; and (3) when $m{<}n$, class-level planning utilizes all experts effectively, yielding consistent gains across rare categories.

%%%%%%%%%%%%%%%---------------------------------%%%%%%%%%%%%%%%%%%%%

\vspace{-5pt}

\subsection{Qualitative Analysis} 
\begin{figure*}
    \centering
    \includegraphics[width=0.92\linewidth]{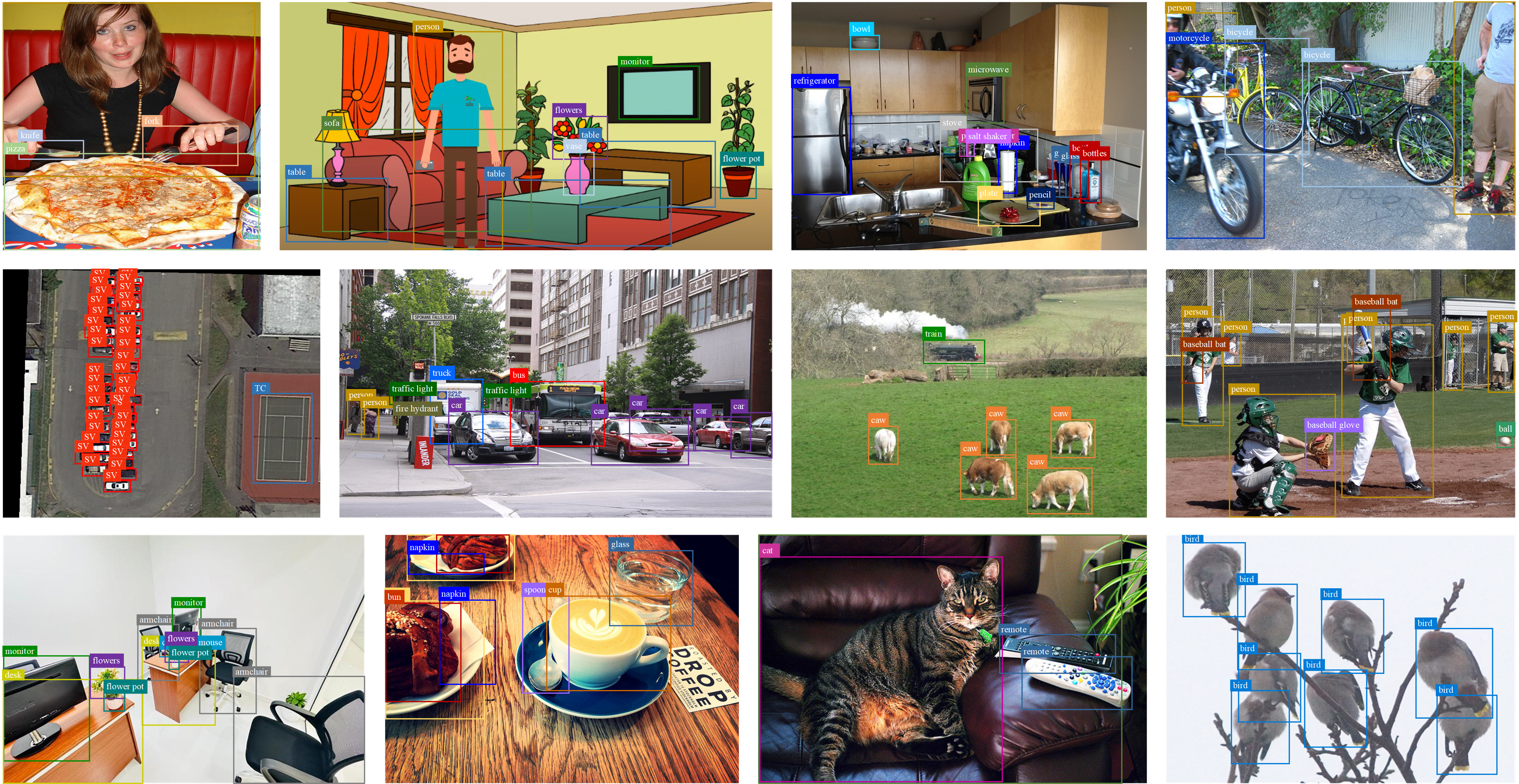}
    \caption{Visualization of some detections using our proposed global expert matching.}
    \label{abl2}
\end{figure*}

Fig.~\ref{attention} presents expert activation maps generated by different routing strategies. A well-specialized model should produce sharp, localized activations focused on distinct objects or regions. In contrast, mode collapse often leads to diffuse or overlapping responses, where multiple experts activate similarly across inputs \cite{chi2022representation}. SoftMoE (column 2), forced by its balancing loss to route all domains to all experts, produces unfocused, diffuse maps, showing no clear specialization where experts respond to unrelated regions. Sinkhorn \cite{clark2022unified} (column 3) uses optimal transport to enforce a uniform distribution, which creates sparser, more distinct activations but fails to align them with specific objects.  GEM without rounding uses the LP planner but omits the compiler. This ignores capacity constraints, thus, overloads a single expert with multiple, unrelated datasets. This creates a confused specialist, resulting in the overlapping and imprecise activations. Our full GEM (column 5) displays an object-aligned attention, indicating clear expert specialization and accurate localization. Fig. \ref{abl2} further shows the detection performance of our proposed GEM-DINO. 

%%%%%%%%%%%%%%%---------------------------------%%%%%%%%%%%%%%%%%%%%

%%%%%%%%%%%%%%%---------------------------------%%%%%%%%%%%%%%%%%%%%

\vspace{-5pt}

\subsection{Ablation}
\subsubsection{Varying Number of Datasets and Experts}
\begin{figure}
    \centering
    \begin{minipage}{0.5\textwidth} 
    \includegraphics[width=0.9\textwidth]{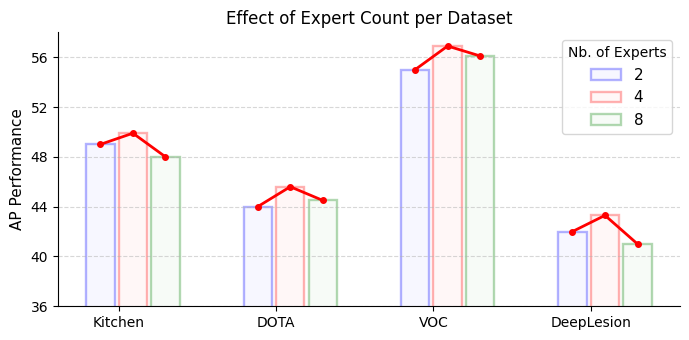}
    \caption{Performance of diverse datasets as experts vary.}   \label{count} 
    \end{minipage}
    \begin{minipage}{0.48\textwidth} 
    \centering
   \includegraphics[width=0.65\textwidth]{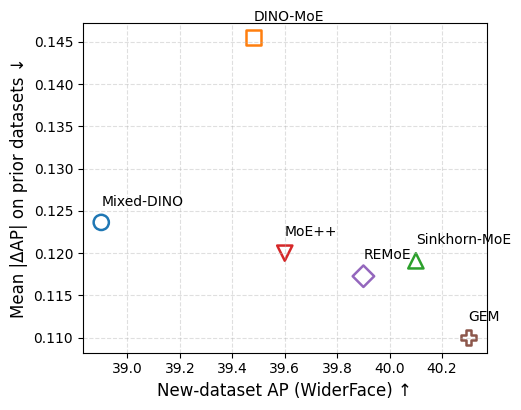}
   \caption{The trade-off between new-task performance (WiderFace AP, x-axis, higher is better) and base-task stability (mean $|\Delta\text{AP}|$ on prior datasets, y-axis, lower is better). Our GEM (bottom-right) achieves the highest adaptability and the lowest forgetting.}
   \label{newdata}
     \end{minipage}
\end{figure}

We first analyze how the number of experts $n$ affects both performance and weight utilization under a fixed total parameter budget. Fig. \ref{count} presents this experiment on a mixture of four diverse datasets ($m=4$: Kitchen, DOTA, DeepLesion, and VOC). While prior MoE works \cite{liu2022sparsity, krajewski2024scaling, clark2022unified} assume that increasing the number of experts linearly improves performance, our results show that this assumption does not hold under parameter constraints. At $n=2$ (overloaded experts), to maintain the budget, each of the two experts is allocated larger weights (2.0x size). However, the GEM assignment ($c_j\!=\!2$) forces it to pack multiple datasets onto each expert. We observe this causes gradient conflict, limiting specialization and resulting in the worst performance. At $n=4$, GEM achieves the ideal configuration, assigning each of the 4 datasets to its own dedicated, large expert (1.0x size). This avoids all task interference and achieves the best performance.
%%%%%%%%%%%%%%%---------------------------------%%%%%%%%%%%%%%%%%%%%
%%%%%%%%%%%%%%%---------------------------------%%%%%%%%%%%%%%%%%%%%
\begin{table*}
\centering
  \begin{minipage}{0.43\textwidth}
 \begin{tabular}{@{}l|cc|cc@{}}\toprule
   &  \multicolumn{2}{c|}{B/32} &\multicolumn{2}{c}{B/16} \\  \cmidrule{2-5}
Routing & JFT & ImageNet & JFT & ImageNet   \\  \midrule
SoftMoE (TopK) \cite{riquelme2021scaling}  & 44.7&  66.2 & 48.6 & 71.5\\
FGMoE \cite{krajewski2024scaling} & 45.3 & 66.4 & 49.3 & 71.6 \\   
Sinkhorn-MoE \cite{liu2022sparsity} & 46.1 & 66.9 & 50.2 & 72.8\\  
GEM W/O Rounding & 45.5 & 67.0 & 50.7 & 72.4 \\
\rowcolor{teal!10} 
GEM (ours)     & 46.9 & 67.4 &51.3  &73.2  \\ \bottomrule
\end{tabular}
    \end{minipage}
 %  \quad
    \begin{minipage}{0.35\textwidth}
        \includegraphics[width=\linewidth]{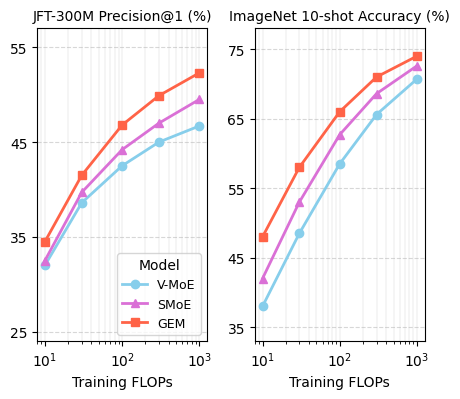}
  %    \caption{}
  %      \label{nb-experts}
    \end{minipage}
     %\hfill
       \caption{Routing ablation and scaling. (Left) Accuracy performance (\%, higher is better) for different routing on ViT-B/32 and ViT-B/16 backbones, evaluated on JFT and ImageNet. GEM outperforms prior MoE routers (Top-K, FGMoE, Sinkhorn) and improves further with rounding. (Right) GEM achieves higher precision/accuracy across a range of training FLOPs on both datasets. }
    \label{abl}
\end{table*}
%%%%%%%%%%%%%%%---------------------------------%%%%%%%%%%%%%%%%%%%%
%%%%
%%%%%%%%%%%%%%%---------------------------------%%%%%%%%%%%%%%%%%%%%% 

Increasing the number of experts further ($n=8$) introduces over-specialization. In this case, the budget is split across 8 smaller experts (each 0.5x size). While one could assign each dataset to multiple small experts (e.g., top-2 fractional matches), this would reintroduce redundant supervision and feature overlap across experts. We assign each dataset to its own expert, leaving some experts inactive; this leads to lower accuracy compared to the balanced $n=4$ configuration. These results demonstrate that simply adding more experts does not guarantee better performance.

For the case where the number of experts exceeds the number of datasets $(m<n)$, we extend GEM to class-level routing in Sec. \ref{class-level}. Furthermore, Fig. \ref{newdata} illustrates the  trade-off between few-shot adaptability (x-axis: new-dataset AP, higher is better) and stability (y-axis: $|\Delta\text{AP}|$ on prior datasets, lower is better). The results show that DINO-MoE suffers from severe forgetting (the highest $\Delta$AP). This occurs because softmax gating reuses the same experts across datasets, causing their parameters to be overwritten by gradients from the new task. GEM achieves both the highest adaptability and the lowest forgetting. As our planner assigns the new task to a dedicated expert, isolating its parameters from the parameters of prior-tasks, thus preventing gradient interference.

%%%%%%%%%%%%%%%---------------------------------%%%%%%%%%%%%%%%%%%%%

\vspace{-2pt}

\subsubsection{Ablation 2: Scale up to large models}
We evaluate the scalability of the GEM planner by integrating it into a large-scale Vision Mixture-of-Experts (V-MoE) \cite{riquelme2021scaling}, replacing its softmax router with our planner-compiler. V-MoE extends the vision transformer by replacing the feedforward networks in selected transformer blocks with MoE layers,  $\text{V-MoE (x)}\!=\!\sum_{j=1}^ng_j(x)\cdot E_j(x)$, where each token \(x\) is routed to a subset of experts (MLP) based on a softmax gating $g_j(x)=\frac{\exp(w_j^\top x)}{\sum_{l=1}^n \exp(w_l^\top x)}$. In this experiment, GEM is applied at the class-level, treating the $m$ classes of the JFT dataset as the tasks to be assigned. The GEM planner solves the LPR and applies hierarchical rounding to find a deterministic assignment  $\rho$ (represented by the binary matrix $\hat{x}$) that maximizes the total affinity $\sum_{i=1}^m \sum_{j=1}^n w_{ij} \hat{x}_{ij}$. This map $\rho$ is then used for routing every class label in the JFT to a specific expert. We apply this to V-MoE B/32 and B/16, using this fixed map for routing while training on JFT-300M, and then evaluating via 10-shot transfer on ImageNet. Table \ref{abl} compares accuracy against different routing, and GEM boosts accuracy by $\sim$4-6\% on both datasets and architectures. We observe that GEM without rounding performs similarly to Sinkhorn-MoE. This behavior is expected, because both methods operate in a fractional regime where each input is softly distributed across multiple experts. In Sinkhorn, this arises from its transport plan, while in our GEM comes from its fractional solution $x^\star$. 

%%%%%%%%%%%%%%%---------------------------------%%%%%%%%%%%%%%%%%%%%

\vspace{-2pt}

%%%%%%%%%%%%%%%---------------------------------%%%%%%%%%%%%%%%%%%%%

\section{Conclusion and future work}
Unified object detection across diverse datasets  is essential for real-world systems, but it faces significant challenges, such as training imbalance, conflicting supervision from heterogeneous label spaces, and performance degradation on low-resource domains. We find that standard Mixture-of-Experts (MoE) models fail in this setting. Their reliance on local gating and a balancing loss creates a fundamental conflict: it enforces a uniform distribution, which prevents the global, specialized assignment needed for multi-domain learning. We proposed GEM (Global Expert Mapping), which re-frames routing as a global optimization problem. GEM's planner (LPR) and compiler (hierarchical rounding) create a deterministic assignment. This plan-then-compile strategy eliminates the need for a balancing loss, enabling true expert specialization and achieving state-of-the-art accuracy, specially on underrepresented datasets. More broadly, our work demonstrates how optimization and compiler principles can reshape MoE design. We proved GEM's flexibility by showing it can scale from dataset-level assignments to fine-grained, class-level routing. This suggests several directions for future research. A key area is extending our planner to continual learning, to handle unseen tasks without catastrophic forgetting. Applying the GEM to multi-modal settings also presents a valuable direction for further exploration.

%%%%%%%%%%%%%%%---------------------------------%%%%%%%%%%%%%%%%%%%% 

{\footnotesize{
\bibliographystyle{IEEEtran}
\bibliography{Final}
}}
%%%%%%%%%%%%%%%%%%%%%%
\vspace{-0.9cm}
\begin{IEEEbiography}
[{\includegraphics[width=1in,height=1in,clip,keepaspectratio]{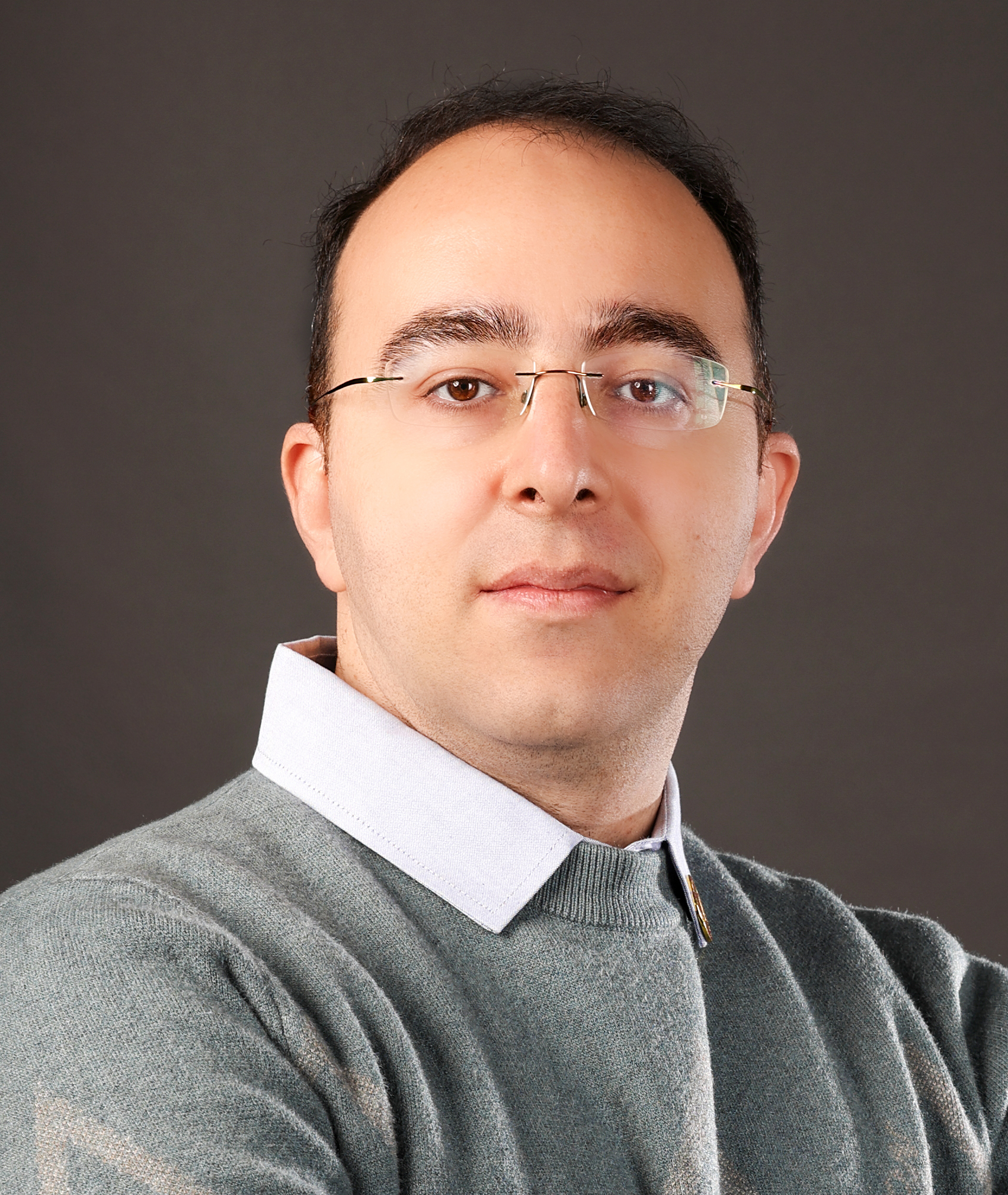}}]
{Pourya Shamsolmoali} (Senior Member, IEEE) received the PhD degree in computer science from Shanghai Jiao Tong University. He has been a visiting researcher with Linköping University, INRIA-France, and ETS-Montreal. He is currently a lecturer with the University of York. His main research focuses on machine learning, computer vision, and image processing.
\end{IEEEbiography}
\vspace{-1.6cm}
%%%%%%%%%%%%%%%%%%%%%%%
\footnotesize{
\begin{IEEEbiography}
[{\includegraphics[width=1in,height=1in,clip,keepaspectratio]{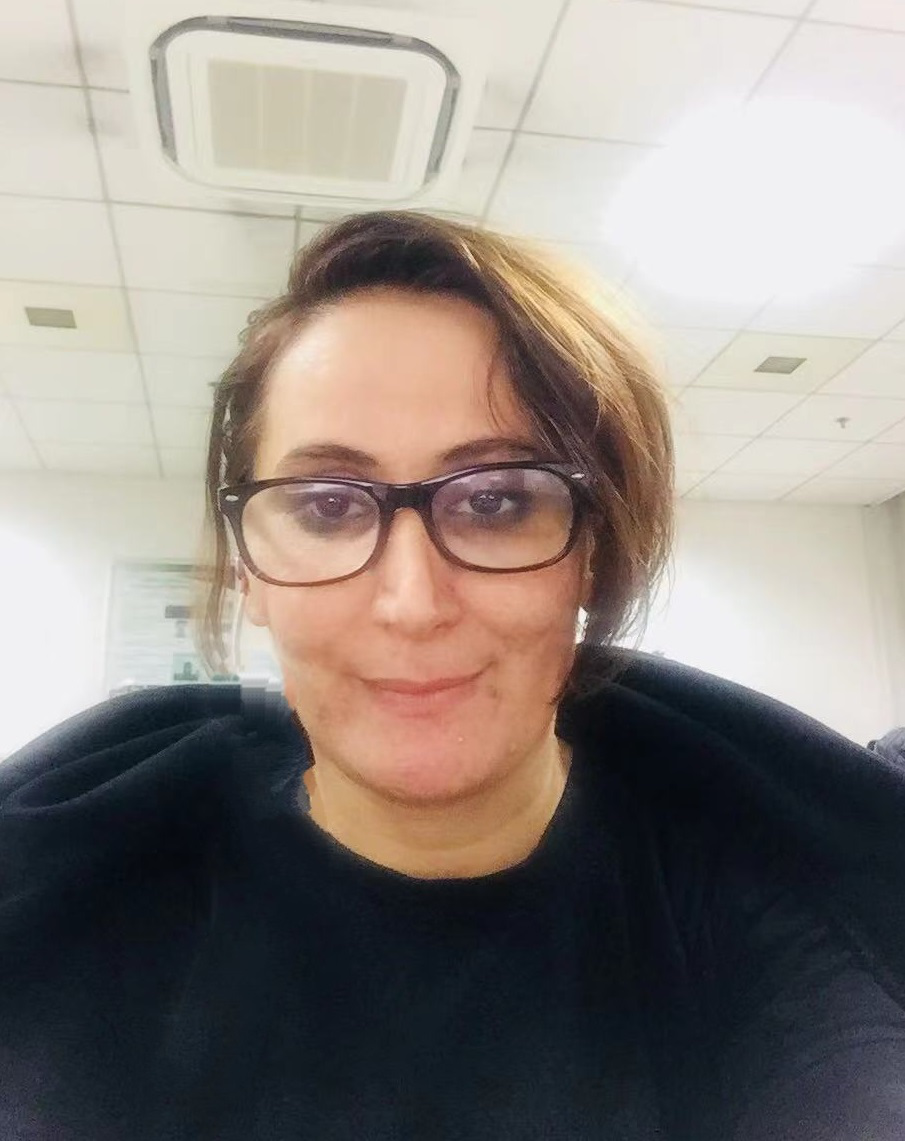}}]
{Masoumeh Zareapoor} (Member, IEEE) received the PhD degree in computer science from Jamia University, India. She is an associate researcher at Shanghai Jiao Tong University and was a research engineer at Huawei, and Northwestern Polytechnical University, Xi'an. Her main research focuses on machine learning and computer vision.
\end{IEEEbiography}}
%%%%%%%%%%%%%%%%%%%%%%%%%%
\vspace{-1.6cm}
%%%%%%%%%%%%%%%%%%%%%%%
\begin{IEEEbiography}
[{\includegraphics[width=1in,height=1in,clip,keepaspectratio]{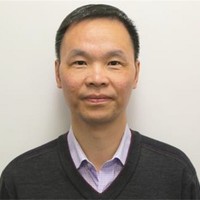}}]
{Huiyu Zhou} received the PhD degree in computer vision from Heriot-Watt University, Edinburgh, U.K. He currently is a full professor with the School of Computing and Mathematical Sciences, University of Leicester, U.K. He has published widely in the field. His research work has been or is being supported by U.K. EPSRC, ESRC, AHRC, MRC, EU, Royal Society, Leverhulme Trust, Puffin Trust, Invest NI and industry.
\end{IEEEbiography}
\vspace{-1.6cm}
%%%%%%%%%%%%%%%%%%%%%
%%%%%%%%%%%%%%%%%%%%
\begin{IEEEbiography}
[{\includegraphics[width=1in,height=1in,clip,keepaspectratio]{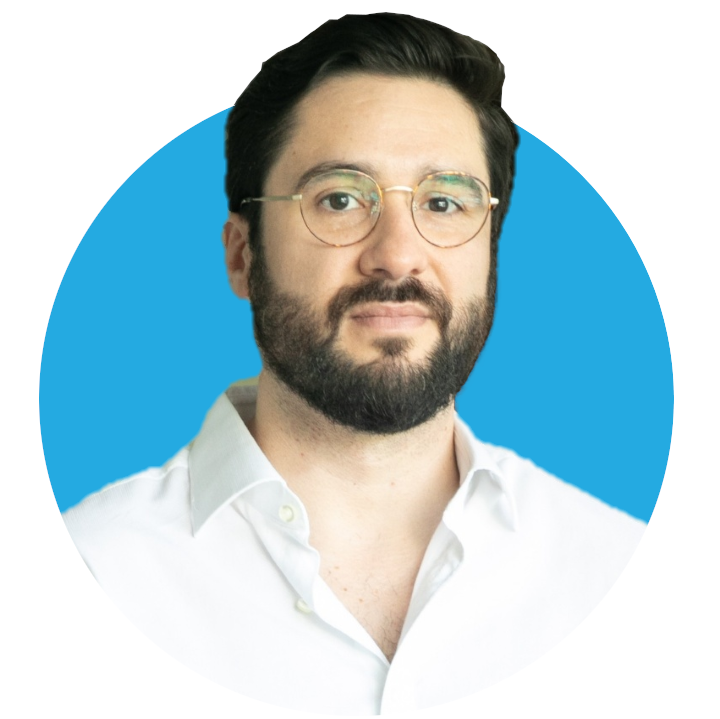}}]
{Oscar Mendez} (Member, IEEE) received the PhD from University of Surrey, U.K. He is currently the Director of AI and Data Science at Locus Robotics, serves as a Visiting Senior Lecturer at the Surrey and a member of the Executive Committee of the British Machine Vision Association. He is an award-winning, internationally recognised researcher, interested in machine learning, robotics and computer vision.
\end{IEEEbiography}
\vspace{-1.6cm}
%%%%%%%%%%%%%%%%%%%%
%%%%%%%%%%%%%%%%%%%%%
\begin{IEEEbiography}
[{\includegraphics[width=1in,height=1in,clip,keepaspectratio]{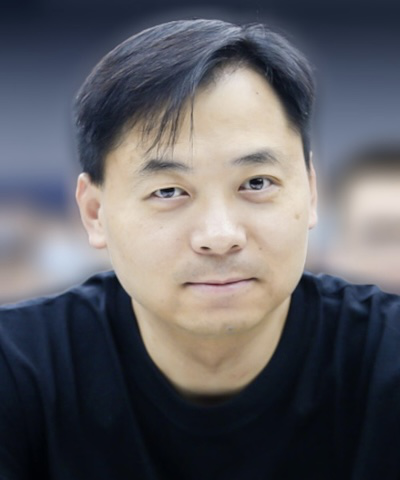}}]
{Dacheng Tao} (Fellow, IEEE) is currently a Distinguished University Professor in the College of Computing \& Data Science at Nanyang Technological University. He mainly applies statistics and mathematics to artificial intelligence, data science, and his research is detailed in one monograph and over 200 publications in prestigious journals and proceedings at leading conferences, with best paper awards, best student paper awards, and test-of-time awards. He received the 2015 and 2020 Australian Eureka Prize, the 2018 IEEE ICDM Research Contributions Award, the 2019 Diploma of The Polish Neural Network Society, and the 2021 IEEE Computer Society McCluskey Technical Achievement Award. He is a Fellow of the Australian Academy of Science, AAAS, ACM and IEEE. 
\end{IEEEbiography}
\vspace{-15cm}
%%%%%%%%%%%%%%%%%%%%%%
\begin{IEEEbiography}
%[{}]
{Xuelong Li} (Fellow, IEEE) is the CTO and Chief Scientist of China Telecom, where he founded the Institute of Artificial Intelligence (TeleAI) of China Telecom.
\end{IEEEbiography}

\end{document}